\pdfoutput=1
\documentclass{article}
\usepackage{iclr2027_conference,times}

\usepackage{amsmath,amsfonts,bm}

\def\eqref#1{equation~\ref{#1}}

\def\1{\bm{1}}

\DeclareMathAlphabet{\mathsfit}{\encodingdefault}{\sfdefault}{m}{sl}
\SetMathAlphabet{\mathsfit}{bold}{\encodingdefault}{\sfdefault}{bx}{n}

\usepackage{hyperref}
\usepackage{url}
\usepackage{booktabs}
\usepackage{graphicx}
\usepackage{xcolor}
\usepackage{amsmath}
\usepackage{array}
\usepackage{float}
\usepackage{placeins}
\usepackage{wrapfig}
\input{numbers.tex}
\input{numbers_a4.tex}
\input{numbers_a67.tex}
\expandafter\def\csname pn@rep.harness.both\endcsname{12}
\expandafter\def\csname pn@rep.harness.opp\endcsname{7}
\expandafter\def\csname pn@rep.harness.p\endcsname{.77}
\expandafter\def\csname pn@rep.harness.pclust\endcsname{.821}
\expandafter\def\csname pn@rep.harness.same\endcsname{5}
\expandafter\def\csname pn@rep.harness.samepct\endcsname{12}
\expandafter\def\csname pn@rep.harness.split1\endcsname{41}
\expandafter\def\csname pn@rep.harness.tasks\endcsname{16}
\expandafter\def\csname pn@rep.harness.units\endcsname{21}
\expandafter\def\csname pn@rep.model.both\endcsname{15}
\expandafter\def\csname pn@rep.model.opp\endcsname{5}
\expandafter\def\csname pn@rep.model.p\endcsname{.302}
\expandafter\def\csname pn@rep.model.pclust\endcsname{.314}
\expandafter\def\csname pn@rep.model.same\endcsname{10}
\expandafter\def\csname pn@rep.model.samepct\endcsname{34}
\expandafter\def\csname pn@rep.model.split1\endcsname{29}
\expandafter\def\csname pn@rep.model.tasks\endcsname{16}
\expandafter\def\csname pn@rep.model.units\endcsname{16}

\input{numbers_ctx.tex}
\expandafter\def\csname pn@ovf.qwen36.hard45.mini.elsewhere\endcsname{0}
\expandafter\def\csname pn@ovf.qwen36.hard45.mini.elsewhere.pp\endcsname{0.0}
\expandafter\def\csname pn@ovf.qwen36.hard45.mini.fail\endcsname{2}
\expandafter\def\csname pn@ovf.qwen36.hard45.mini.n\endcsname{3}
\expandafter\def\csname pn@ovf.qwen36.pool447.mini.elsewhere\endcsname{1}
\expandafter\def\csname pn@ovf.qwen36.pool447.mini.elsewhere.pp\endcsname{0.2}
\expandafter\def\csname pn@ovf.qwen36.pool447.mini.fail\endcsname{6}
\expandafter\def\csname pn@ovf.qwen36.pool447.mini.n\endcsname{6}
\expandafter\def\csname pn@ovf.qwen38.hard45.mini.elsewhere\endcsname{3}
\expandafter\def\csname pn@ovf.qwen38.hard45.mini.elsewhere.pp\endcsname{6.7}
\expandafter\def\csname pn@ovf.qwen38.hard45.mini.fail\endcsname{11}
\expandafter\def\csname pn@ovf.qwen38.hard45.mini.n\endcsname{18}
\expandafter\def\csname pn@ovf.qwen38.hard45.mini.run2.again\endcsname{7}
\expandafter\def\csname pn@ovf.qwen38.hard45.mini.run2.passed\endcsname{3}
\expandafter\def\csname pn@ovf.qwen38.hard45.mini.run2.present\endcsname{11}
\expandafter\def\csname pn@ovf.qwen38.pool447.mini.elsewhere\endcsname{6}
\expandafter\def\csname pn@ovf.qwen38.pool447.mini.elsewhere.pp\endcsname{1.3}
\expandafter\def\csname pn@ovf.qwen38.pool447.mini.fail\endcsname{18}
\expandafter\def\csname pn@ovf.qwen38.pool447.mini.n\endcsname{34}

\title{What Does a Harness Buy? \\ Tokens, Mostly.}

\author{Yangze Liu \\
Shandong University \\
\texttt{yangze2@illinois.edu}
\And
Zhongyi Han\thanks{Corresponding author.} \\
Shandong University \\
\texttt{zhongyi.han@sdu.edu.cn}}

\iclrfinalcopy

\newcommand{\rev}[1]{{#1}}
\newcommand{\fitbox}[1]{\sbox0{#1}\ifdim\wd0>\linewidth\resizebox{\linewidth}{!}{\usebox0}\else\usebox0\fi}

\newcommand{\capfigone}{\rev{\textbf{Same model, different harness: no harness leads consistently on hard45.} (a) Input tokens per task against pass rate on hard45 for seven models in three harnesses. Each point is one configuration, averaged over its runs. Shape marks the harness and colour the model. For points of one colour, the vertical spread is the difference in pass rate between harnesses and the horizontal spread is the difference in tokens. (b) Difference in pass rate for each pair of harnesses on hard45, one point per model; a two-coloured point is two models with the same difference. The diamond is the mean over the seven models with its 95\% interval. All three diamonds cover zero, and single models fall on both sides.}}

\newcommand{\captabresults}{\rev{\textbf{No harness has the highest pass rate on every model.} Tasks passed out of 45 on hard45, and pass rate (\%) on the other task sets, for each model in the three harnesses. Each number is one run, so an entry with several numbers lists a run and its reruns. A denominator is shown where trials lost to infrastructure left fewer than 45 tasks. The runs were made in two rounds, and results from the second round are in bold.}}

\newcommand{\captabpool}{\rev{\textbf{On pool447 eight of twelve differences between two harnesses are bounded within 5 points, and the average difference between Claude Code and mini-SWE-agent within 2.} Difference in pass rate (first harness minus second, in points) with its 90\% interval, for each model and each pair of harnesses, over the tasks scored in both, so a value can differ slightly from a subtraction in Table~\ref{tab:results}. The Qwen rows use each harness's first run. ``Within'' names the smallest of $\pm$2, $\pm$3, and $\pm$5 points that contains the whole interval, which excludes a larger difference in either direction, and a dash marks an interval that reaches beyond $\pm$5. The average weights the four models equally and bounds the average over models only. The last row tests whether the difference is the same on all four models (Cochran's $Q$), and the average is given for the one pair on which this test does not reject. $^{\dagger}$The 391 tasks of Table~\ref{tab:results}; the average takes this model on the 404 tasks that Claude Code and mini-SWE-agent both completed ($+$2.2).}}

\newcommand{\capfignoise}{\textbf{A harness swap flips about as many tasks as a rerun, and a model swap flips more.} Each point is one pair of runs on the same task set, and its value is the flip rate, the share of tasks whose outcome differs between the two runs. Rerun pairs share model and harness, harness-swap pairs share the model, and model-swap pairs share the harness. \rev{Black points are hard45 over all models (22 rerun pairs, 21 harness swaps, 70 model swaps), with a bar at their median. Diamonds are pool447, coloured by model.}}

\newcommand{\captabconditions}{\rev{\textbf{Conditions that were not equal across harnesses, and their checks: only the output cap is tied to a loss of tasks.} Each row gives one condition, the check we ran, and its result. Where three pass counts or pass rates are listed, the order is Claude Code, mini-SWE-agent, OpenCode.}}

\newcommand{\captabcost}{\rev{\textbf{How much more one harness costs than another depends on the model, from under 10\% to 3 to 4 times.} Mean cost per trial in each harness. Two values in an entry are the first run and its rerun. The DeepSeek-V4.1-Flash rows pool two runs, and the other single values are first runs. Costs are comparable within a row only. The last two columns divide the most expensive harness by the cheapest, per trial and per solved task, the latter pooled over the runs shown. DeepSeek-V4.1-Flash is priced at the list prices of DeepSeek-V4-Flash, and Claude Opus 5.5 at Anthropic's list prices.}}

\newcommand{\capfigctx}{\rev{\textbf{Input per step and number of steps weigh differently on the two models.}} Input tokens of each model call against the step number, for GLM-5.3-Flash and DeepSeek-V4-Flash on hard45. Each faint line is one trial, and the bold line is the median over the trials still running at that step, drawn while at least half of them are. \rev{The height at which a line starts is the preamble, its slope is what each step adds to the history, and its length is the number of steps.} Sharp drops are compactions of the history.}

\begin{document}

\maketitle
\lhead{Preprint.}

\begin{abstract}
A coding agent is a language model inside a harness: the system prompt, the tools, and the context management that let a chat model work in a repository.
Most of the engineering on these agents goes into the harness, and vendors and leaderboards credit it with score gains.
\rev{Earlier studies have fixed the model and swapped the harness, but have not measured whether the swap changes the pass rate more than running the same harness again.}
\rev{We run seven models through three production harnesses, Claude Code, mini-SWE-agent, and OpenCode, on SWE-bench Verified and two benchmarks of other kinds, and rerun them unchanged.} %
\rev{No harness leads consistently across models.}
\rev{Where the data can bound the difference between two harnesses, the bound is 2 to 5 points, and a harness swap changes about as many task outcomes as a rerun.} %
\rev{The exceptions are harness effects in particular pairings of model and harness, whose direction changes with the model: OpenCode trails by 8 to 9 points on one model and leads by 14 on another.} %
\rev{Cost differences between harnesses also depend on the model, from under 10\% to 3 to 4 times.} %
\rev{Which harness is better therefore has no answer apart from the model, and a harness difference should be reported next to a rerun.}
\end{abstract}

\section{Introduction}

Coding agents have turned into a product category, and the product is a model plus a harness.
The harness, which part of the literature calls a scaffold, is everything around the model: the system prompt, the tools it can call, the way tool output is fed back, and the way the conversation is trimmed when it grows too long.
It is also the part that changes most: popular harnesses ship more than two releases a day \citep{bensghaier2026dontblame}, and most of what changes is prompts, tools, and context management.

The field has come to believe that this engineering moves the score, and the belief has evidence behind it.
On a public terminal benchmark, the same model is listed at 52.1\% under one harness and 57.8\% under another, and the lower-scoring harness consumes 65$\times$ the input tokens \citep{vats2026scaffoldeffect}.
Six harnesses run over the same model pool span 23.8 points \citep{yao2026harnessbench}, and five harnesses on one fixed model span 27.4 points \citep{zheng2026clawswebench}.
Vendors sell harnesses on this basis: one advertises a 2.4-point gain over Claude Code with its own model held fixed \citep{zhipu2026zcode}.
Leaderboards, meanwhile, mix harnesses freely, so a ranking of models is also a ranking of whatever harness each submitter chose \citep{zhang2026stopcomparing}.

These numbers are hard to read, for two reasons.
First, almost all of them come from a single run of each model in each harness, while ten repeated runs of one unchanged setup on SWE-bench Verified \citep{jimenez2024swebench,openai2024verified} already spread by up to 6 points \citep{bjarnason2026randomness}.
Second, the harnesses being compared are often not on the same footing.
In the five-harness study above, one harness scores 19.1\% with a minimal adapter to the benchmark and 73.4\% with the full adapter \citep{zheng2026clawswebench}, so part of what is measured is how well each harness was connected to the benchmark.
The study closest to ours runs three production harnesses on two models over 50 tasks, once each, and finds differences of 0 to 8 points within a model, with intervals that cover zero for all but the largest \citep{vats2026scaffoldeffect}.
Whether those 8 points come from the harness or from repeating the run cannot be told from that design, since repeated runs alone spread nearly as far.
\rev{A recent benchmark, Agents' Last Exam \citep{sun2026ale}, reports that five harnesses span 4.9 points with the model fixed while twelve models span 16.8 points with the harness fixed, again mostly from single runs with no repeated run to compare against.} %
So the question a harness claim rests on, whether swapping the harness moves a score more than running the same harness again, has not been measured.

We measure it by using the rerun as the yardstick.
A \emph{rerun} is the same model in the same harness on the same tasks, run again with nothing changed.
We take three production harnesses as they ship, Claude Code, mini-SWE-agent, and OpenCode \citep{claudecode,minisweagent,opencode}, and run \rev{seven models} through all three on SWE-bench Verified \citep{jimenez2024swebench,openai2024verified}: Qwen3.6-35B-A3B, Qwen3.8-27B, DeepSeek-V4-Flash, \rev{DeepSeek-V4.1-Flash}, GLM-5.3-Flash, HY4-Preview, \rev{and Claude Opus 5.5}. %
\rev{Every model is rerun, five of the seven in all three harnesses, so a difference between two harnesses can be read next to the difference between two runs of one harness.} %
Two disjoint task sets play different roles.
\emph{hard45} is the 45 tasks in the two hardest annotated difficulty bands, where we expect a harness to matter most.
\emph{pool447} is the remaining 447 tasks that run in our setup, 445 of which can be scored, and it is large enough to bound a small difference. %
\rev{Two benchmarks of other kinds, spreadsheet editing and data analysis, test whether the picture is specific to repository repair.}
All runs are offline: the agent can reach the model and nothing else on the network.

\begin{figure}[t]
\centering
\includegraphics[width=\linewidth]{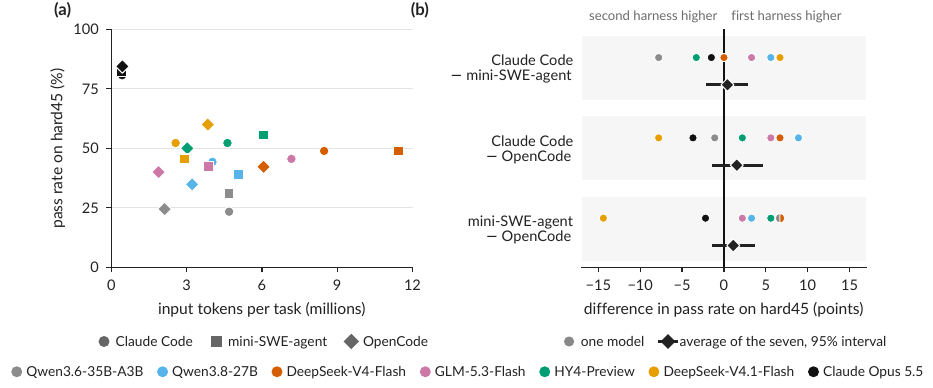}
\caption{\capfigone}
\label{fig:fig1}
\end{figure}

\rev{No harness leads consistently across the seven models (Figure~\ref{fig:fig1}).}
\rev{On hard45 each of the three harnesses has the highest pass rate on some model, and averaged over the seven models no pair of harnesses differs.} %
\rev{On pool447, where the tasks suffice for a bound, eight of the twelve differences between two harnesses on a single model are within 5 points, and the difference between Claude Code and mini-SWE-agent averaged over models is within 2.} %
A harness swap also changes the outcome of about as many tasks as a rerun does: on hard45 both change \pn{flip.hard45.rerun.median}\% of tasks at the median, and the tasks that one harness alone passes in one run are not the tasks it alone passes in the next.

\rev{Harness effects on pass rate do exist, in particular pairings of model and harness, and their direction changes with the model.}
\rev{With Qwen3.8-27B on pool447, OpenCode trails Claude Code and mini-SWE-agent by 8 to 9 points, a deficit that goes with model responses cut off at the output cap, which OpenCode does not continue.} %
\rev{With DeepSeek-V4.1-Flash on hard45, the same harness leads mini-SWE-agent by 14 points over two runs.} %
\rev{A harness difference measured on one model therefore does not carry over to another.}

\rev{Cost differences between harnesses depend on the model as well.}
\rev{For the same model on the same tasks, the most expensive harness costs under 10\% more than the cheapest on some models and 3 to 4 times as much on another, with cost computed from each run's own tokens at the vendor's list price.} %
\rev{The differences come from three things: the preamble of system prompt and tool definitions that a harness sends with every step, what it keeps in the conversation history, and how many steps the model takes.}
\rev{Which of the three dominates depends on the model, so the harness that sends the least per step can end up consuming the most tokens.}

For anyone comparing harnesses, the rerun is the missing control.
\rev{With 45 tasks, a difference of a dozen points between two harnesses is detected only about half the time, and with 445 tasks a difference of about \pn{mde.pool447.pp} points is detected four times in five.} %
Many published harness gains are smaller than the first of these sizes, and some are smaller than the second.

\section{Related Work}

Earlier work has compared harnesses under a fixed model and has measured how much repeated runs vary, each without the other.

\paragraph{Harness comparisons.}
Several recent studies swap whole harnesses under a fixed model and report large spreads.
Beyond the two spreads cited in the introduction \citep{yao2026harnessbench,zheng2026clawswebench}, one fixed model costs 139$\times$ more tokens under its worst harness than under its best on a single git task \citep{alier2026scaffolding}.
Two results in this line anticipate ours.
Three production harnesses paired on the same terminal tasks differ by 0 to 8 points with intervals that cover zero for all but the largest, while tokens per solved task differ 40$\times$ \citep{vats2026scaffoldeffect}, and 35 consecutive releases of one harness bring no significant score gain on a fixed model while nearly doubling token use \citep{bensghaier2026dontblame}.
The earlier harness papers on SWE-bench each compared their own design against the others under a fixed model \citep{yang2024sweagent,wang2024openhands,xia2024agentless}.
\rev{Agents' Last Exam \citep{sun2026ale} fixes a model and varies the harness in its main table, and it points the same way as our results: with one model fixed, five harnesses span 4.9 points, and with one harness fixed, twelve models span 16.8.} %
\rev{Its design goal is to rank systems, so the groups that swap the harness are mostly single runs, and it reports no intervals, tests, or rerun reference.} %
\rev{In the group that fixes Claude Opus 4.7 across three harnesses, for example, the spread is 7.2 points and no combination is run twice.} %
\rev{How much of 4.9 or 7.2 points belongs to the harness and how much to repeating the run is the question that design leaves open.} %
Every study in this line runs each combination of model and harness once or twice.
Rankings also move with the harness: which model is most robust to meaning-preserving rewrites of a task depends on whether it runs under mini-SWE-agent or OpenCode \citep{mahmud2026jagged}, and a position paper asks that comparisons disclose the harness for this reason \citep{zhang2026stopcomparing}.
We keep the whole harness as the unit and the production software as shipped, and add the two things this line lacks: reruns of the same combination, and cost computed from one price list per model.

\paragraph{What is inside a harness.}
A source-code taxonomy of coding agents lists twelve design dimensions and asks for the control we use, a fixed model with the harness varied \citep{rombaut2026inside}.
Our three harnesses sit at different points of that taxonomy: mini-SWE-agent gives the model a single \texttt{bash} tool and no context management \citep{yang2024sweagent,minisweagent}, while Claude Code and OpenCode add typed file tools, long system prompts, and automatic compaction \citep{claudecode,opencode}.
Component studies show that individual pieces matter under pressure: a context-management package nearly doubles a small model's score in a 20k-token window \citep{lewis2026samemodel}, and restricting the tools changes cost by task regime and agent while pass rates stay tied \citep{yang2026executecode}.
A component study on SWE-bench Verified varies context management, action space, and planning inside one lightweight harness, and finds that capable models do as well with a bash-only tool set at lower cost \citep{fan2026harnessdesign}.
We ask the complementary question: what the whole production harness, as users run it, changes against a rerun of itself.
Models fine-tuned on one harness's trajectories lose as much as 29 points when moved to another, while base models vary far less \citep{thangarajah2026dcas}.
Our check of DeepSeek's own harness with DeepSeek-V4-Flash asks the same question of a model served through an API.

\paragraph{Noise and power in agent evaluation.}
The case for error bars on evaluations has been made for language models in general \citep{miller2024errorbars} and for agents in particular, where a score reported without its cost or its variance rewards the wrong things \citep{kapoor2024agents}.
Ten reruns of one model in one harness on SWE-bench Verified spread by 2.2 to 6.0 points, with standard deviations above 1.5 points even at temperature zero, so reported gains of 2 to 3 points can be noise \citep{bjarnason2026randomness}.
\rev{That study measures rerun variation without swapping the harness.} %
A generalizability-theory decomposition finds that the agent main effect explains under 3\% of outcome variance while agent-by-task interaction explains 7 to 23\% \citep{srinivasan2026reliability}, and a pre-registered comparison on GAIA fits the harness-by-model interaction directly and finds it significant \citep{starace2026gaia}.
\rev{The effects we find in particular pairings of model and harness are that interaction, observed with production harnesses on coding tasks.}
We take rerun variation as the yardstick and report each harness difference next to it, together with the smallest difference our task sets can detect.

\section{Setup}

The design fixes the model and the tasks, swaps the harness, and repeats runs with nothing changed, so that every harness difference has a rerun to compare against.

\paragraph{Harnesses.}
We run three production harnesses as they ship, each at pinned versions: Claude Code 2.1 (releases 2.1.259 to 2.1.273; 2.1.263 with GLM-5.3-Flash and Claude Opus 5; \rev{2.1.280 with Claude Opus 5.5}), mini-SWE-agent 2.4.6, and OpenCode 1.18 (1.18.30 and 1.18.31) \citep{claudecode,minisweagent,opencode}. %
They cover one minimal harness and two full products, one closed and one open.
mini-SWE-agent gives the model a single \texttt{bash} tool and has no context management, so a trial, one attempt at one task, stops when the conversation history exceeds the model's context window.
Claude Code and OpenCode give the model typed file, search, and edit tools, prepend long system prompts, and compact the history automatically when it grows.
None of the three sends sampling parameters, so decoding is the provider's default.
\rev{A step is one model call, and the step cap is listed with the other settings in Appendix~\ref{app:controls}.}
\rev{Wall-clock limits, the handling of the output cap, and the interface through which Claude Code reaches the model were not the same for all three harnesses, and Table~\ref{tab:conditions} in Appendix~\ref{app:checks} lists each such condition with the check we ran for it.}
Where the model's interface accepts a reasoning-effort setting, with GLM-5.3-Flash \rev{and Claude Opus 5.5}, all three harnesses set it to \texttt{high}. %
Everything else is left as the harness authors set it (Table~\ref{tab:controls} in the appendix).
We also run DeepSeek's own command-line harness with DeepSeek-V4-Flash as a check (Section~\ref{sec:checks}).

\paragraph{Models.}
\rev{Seven models run in all three harnesses.} %
Qwen3.6-35B-A3B and Qwen3.8-27B are served locally with vLLM 0.27.1 in bfloat16 precision.
DeepSeek-V4-Flash (release 0731), \rev{DeepSeek-V4.1-Flash}, GLM-5.3-Flash, and HY4-Preview are called through their vendors' APIs.
\rev{DeepSeek-V4.1-Flash is the vendor's replacement for DeepSeek-V4-Flash, and we treat it as a separate model.} %
\rev{Claude Opus 5.5 is called through the native Anthropic interface in all three harnesses, whereas for models from other vendors Claude Code reaches the model through a compatibility layer.} %
Claude Opus 5 runs in Claude Code only.
DeepSeek-V4-Flash, HY4-Preview, and GLM-5.3-Flash ranked second to fourth by tokens processed on OpenRouter in the 30 days to 2026-09-20 \citep{openrouter2026rankings}.
Harbor 0.22.0 \citep{harbor} orchestrates the runs, one container per trial, and scores each trial with the benchmark's own tests.

\paragraph{Tasks.}
SWE-bench Verified \citep{jimenez2024swebench,openai2024verified} has 500 tasks.
Eight need a public web service to verify and are dropped, and we divide the remaining 492 into two disjoint task sets.
hard45 is every task in the two hardest annotated difficulty bands, 1 to 4 hours and over 4 hours (42 and 3 tasks), taken whole so that no sampling is involved.
pool447 is the other 447 tasks, of which 445 can be scored. %
\rev{All seven models run on hard45, and four of them also run on pool447: the two Qwen models, DeepSeek-V4.1-Flash, and Claude Opus 5.5.} %
\rev{For other kinds of work we add SpreadsheetBench Verified, 400 spreadsheet-editing tasks, and 150 tasks drawn at random from the hard tasks of DABstep, a data-analysis benchmark, both scored deterministically and both run with DeepSeek-V4.1-Flash.} %
During a trial the agent has no network access beyond the model endpoint.
In early runs with web access, 71\% of DeepSeek and Qwen trials fetched the upstream fix from GitHub, so those runs are discarded and the offline condition is enforced at the network layer (Appendix~\ref{app:leakage}).
OpenCode's search tools download a binary at first use, which fails offline, so we install the binary in advance (Appendix~\ref{app:sensitivity} reports OpenCode as shipped).

\paragraph{Terms.}
A \emph{task} is one benchmark problem, and a \emph{task set} is hard45, pool447, or one of the two other benchmarks.
A \emph{configuration} is one model in one harness on one task set with one fixed set of settings.
A \emph{trial} is one attempt by a configuration at one task, and a \emph{run} is a configuration attempting its whole task set once, one trial per task.
A \emph{rerun} is another run of the same configuration with nothing changed.
A \emph{harness swap} compares two runs of the same model on the same task set under different harnesses, and a model swap compares two runs under the same harness with different models.
A \emph{flip} is a task whose outcome, pass or fail, differs between two runs, and the \emph{flip rate} of two runs is the share of such tasks.
\emph{Points} are percentage points.
A \emph{difference} always refers to pass rate, and costs are compared as multiples or percentages.

\paragraph{Scoring and statistics.}
A trial that reaches a step, time, context, or output limit is scored like any other, and most such trials fail.
A trial lost to an infrastructure fault leaves the denominator (Appendix~\ref{app:pools}).
Every difference between two harnesses is paired over the same tasks.
When each side has one run, we use McNemar's exact test and a Newcombe interval.
\rev{When a side has several runs, we average each task's outcome over the runs and compute the interval over tasks, by a bootstrap or a paired $t$ interval.}
\rev{On pool447 we report whether the 90\% interval of a difference lies within $\pm$2, $\pm$3, or $\pm$5 points, which excludes a larger difference in either direction.} %
\rev{Corrected $p$-values use Holm's method for multiple comparisons.}
Cost is each trial's own token counts, divided into cached input, uncached input, and output, multiplied by the vendor's list price for the model.
\rev{The two locally served models have token counts and no list price, so cost is reported for the five models called through an API.} %

\section{Results}

\subsection{No harness leads consistently across models}
\label{sec:nolead}

\begin{table}[t]
\centering
\caption{\captabresults}
\label{tab:results}
\fitbox{\footnotesize\setlength{\tabcolsep}{7pt}
\begin{tabular}{@{}lccc@{}}
\toprule
Model & Claude Code & mini-SWE-agent & OpenCode \\
\midrule
\multicolumn{4}{@{}l}{\textit{hard45: tasks passed out of 45}} \\
Qwen3.6-35B-A3B & 11, 10 & 15, 13 & 11, 11 \\
Qwen3.8-27B & 19/43, 20 & 17, 18 & 17, 14/44 \\
DeepSeek-V4-Flash & 21, 23 & 22 & 19 \\
GLM-5.3-Flash & 22, \textbf{19} & 20, \textbf{18} & 18, \textbf{18} \\
HY4-Preview & 23, \textbf{24} & 26, \textbf{24} & 21, \textbf{24} \\
DeepSeek-V4.1-Flash & \textbf{24, 23} & \textbf{21, 20} & \textbf{27, 27} \\
Claude Opus 5.5 & \textbf{36, 37, 36} & \textbf{37} & \textbf{38} \\
Claude Opus 5 & 36, 35, 35 & n/a & n/a \\
\addlinespace
\multicolumn{4}{@{}l}{\textit{pool447: pass rate (\%)}} \\
Qwen3.6-35B-A3B & 69.7 & 71.5, 70.3 & 66.6 \\
Qwen3.8-27B & 78.2 & 79.3 & 70.1 \\
DeepSeek-V4.1-Flash & \textbf{82.7, 84.0} & \textbf{82.7, 80.9} & \textbf{82.9, 84.3} \\
Claude Opus 5.5 (391 tasks)$^{\dagger}$ & \textbf{95.1} & \textbf{93.1} & \textbf{93.4} \\
\addlinespace
\multicolumn{4}{@{}l}{\textit{Two other benchmarks, DeepSeek-V4.1-Flash: pass rate (\%)}} \\
SpreadsheetBench Verified & \textbf{88.3, 88.5} & \textbf{88.8, 87.0} & \textbf{88.0, 88.8} \\
DABstep & \textbf{74.0, 74.7} & \textbf{72.0, 76.0} & \textbf{71.3, 75.3} \\
\bottomrule
\end{tabular}
}

\vspace{2pt}
\parbox{0.86\linewidth}{\footnotesize\rev{$^{\dagger}$Tasks ran in a pre-drawn random order, and the row gives the 391 tasks that all three harnesses completed. Claude Code completed all 447 and passed 422.}}
\end{table}

\rev{No harness leads consistently across the seven models.}
\rev{On hard45, the harness with the highest pass rate changes from model to model, and averaged over the models no pair of harnesses differs (Table~\ref{tab:results}).}
\rev{Each harness is the top one on two models, and on the seventh, DeepSeek-V4-Flash, Claude Code and mini-SWE-agent tie.} %
\rev{Averaged over the seven models on hard45, Claude Code is $0.4$ points above mini-SWE-agent and $1.5$ above OpenCode, mini-SWE-agent is $1.1$ above OpenCode, and all three 95\% intervals cover zero (Figure~\ref{fig:fig1}b, Table~\ref{tab:pairs}).} %
\rev{Behind these averages, single-model differences point both ways, from $-14$ to $+9$ points, and Section~\ref{sec:losses} examines the largest.} %
\rev{A ranking of harnesses read from one model is reversed by another.}

\begin{table}[t]
\centering
\caption{\captabpool}
\label{tab:pool447}
\fitbox{\footnotesize\setlength{\tabcolsep}{3.2pt}
\begin{tabular}{@{}l rcc rcc rcc@{}}
\toprule
 & \multicolumn{3}{c}{Claude Code $-$ mini-SWE-agent} & \multicolumn{3}{c}{Claude Code $-$ OpenCode} & \multicolumn{3}{c}{mini-SWE-agent $-$ OpenCode} \\
\cmidrule(lr){2-4}\cmidrule(lr){5-7}\cmidrule(l){8-10}
Model & diff. & 90\% interval & within & diff. & 90\% interval & within & diff. & 90\% interval & within \\
\midrule
Qwen3.6-35B-A3B & $-$1.8 & [$-$4.7, +1.1] & $\pm$5 & +2.9 & [$-$0.3, +6.1] & -- & +4.7 & [+1.6, +7.8] & -- \\
Qwen3.8-27B & $-$1.4 & [$-$3.7, +1.0] & $\pm$5 & +7.9 & [+4.8, +10.9] & -- & +9.2 & [+6.1, +12.3] & -- \\
DeepSeek-V4.1-Flash & +1.6 & [$-$0.0, +3.2] & $\pm$5 & $-$0.2 & [$-$1.8, +1.4] & $\pm$2 & $-$1.8 & [$-$3.2, $-$0.4] & $\pm$5 \\
Claude Opus 5.5$^{\dagger}$ & +2.0 & [+0.4, +3.9] & $\pm$5 & +1.8 & [+0.2, +3.5] & $\pm$5 & $-$0.3 & [$-$1.7, +1.2] & $\pm$2 \\
\midrule
Average of the four & +0.2 & [$-$0.9, +1.2] & $\pm$2 & \multicolumn{3}{c}{not averaged} & \multicolumn{3}{c}{not averaged} \\
Do the models disagree? ($p$) & \multicolumn{3}{c}{0.06} & \multicolumn{3}{c}{0.002} & \multicolumn{3}{c}{$<$0.001} \\
\bottomrule
\end{tabular}
}
\end{table}

\rev{On pool447, the task set large enough to give a bound, eight of the twelve single-model differences between two harnesses are within 5 points (Table~\ref{tab:pool447}).} %
\rev{With DeepSeek-V4.1-Flash on pool447, run twice in every harness, the 90\% intervals of all three pairs exclude a difference beyond 3.2 points.} %
\rev{Averaged over the four models on pool447, Claude Code and mini-SWE-agent, which send the most and the least text with every step, differ by 0.2 points (90\% interval $[-0.9, +1.2]$).} %
\rev{The pairs that are not bounded this way are those with OpenCode on the two Qwen models, and Section~\ref{sec:losses} takes up the largest of them.} %

\rev{On the two Qwen models, point estimates put Claude Code 1.8 and 1.4 points behind mini-SWE-agent, and the upper ends of the 90\% intervals, $+1.1$ and $+1.0$, bound how far it could be ahead.} %
\rev{With 445 tasks, most single-model intervals are too wide to exclude a 2-point difference (Section~\ref{sec:noise}).} %
\rev{The 2-point bound is therefore a statement about the average over models, and the 5-point bound is the one that holds model by model.} %

\rev{The size of a difference read from one run is not stable.}
\rev{With DeepSeek-V4.1-Flash on pool447, the three harnesses were within 0.2 points of one another in the first run.} %
\rev{In the second run mini-SWE-agent was 3.4 points below OpenCode, a difference that is significant within that run (Holm-corrected $p$ = 0.024).} %
\rev{Over both runs the difference is 1.8 points and no pair is significant after correction.} %
\rev{Read alone, the second run would have been reported as a harness effect.}

\rev{On a frontier model reached through its native interface, any lead for Claude Code is small.}
\rev{Claude Opus 5.5 runs natively in all three harnesses, which is the setting most favourable to Claude Code: if the compatibility layer it uses for other vendors' models had held it back, it should lead here.} %
\rev{On hard45 it does not lead: Claude Code passes 36, 37, and 36 tasks in three runs, against 37 for mini-SWE-agent and 38 for OpenCode.} %
\rev{On pool447, over the 391 tasks that all three harnesses completed, Claude Code is about 2 points above each of the other two, a lead that is not significant after Holm correction and that the 90\% intervals place between 0 and 4 points.} %

\subsection{A harness swap changes about as many outcomes as a rerun}
\label{sec:noise}

\begin{figure}[t]
\centering
\begin{minipage}[c]{0.50\linewidth}
\includegraphics[width=\linewidth]{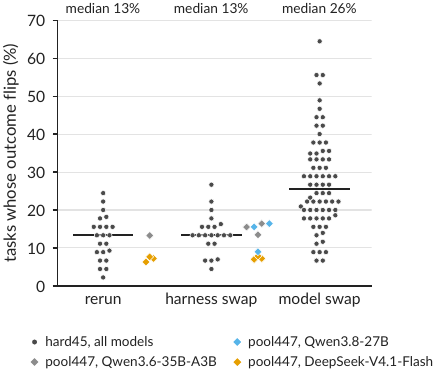}
\end{minipage}\hfill
\begin{minipage}[c]{0.46\linewidth}
\caption{\capfignoise}
\label{fig:noise}
\end{minipage}
\end{figure}

\rev{A harness swap changes the outcome of about as many tasks as a rerun does (Figure~\ref{fig:noise}).}
The flip rate of two runs, the share of tasks that pass in one and fail in the other, puts a rerun, a harness swap, and a model swap on one scale.
On hard45, reruns flip \pn{flip.hard45.rerun.median}\% of tasks at the median and harness swaps the same \pn{flip.hard45.harness.median}\%, while model swaps flip \rev{26}\%. %
\rev{On pool447 with DeepSeek-V4.1-Flash, a harness swap flips 7 to 8\% of tasks and a rerun 6 to 8\%.} %
Pass rates move less than these flip rates because flips go in both directions and mostly cancel.

\rev{Behaviour agrees with the outcomes: after a harness swap the agent edits about the same files as after a rerun.}
\rev{On the five models that have reruns in all three harnesses on hard45, we compare the sets of source files that two trials on the same task end up editing, by Jaccard similarity, where 1 means identical sets.} %
\rev{Two trials from a rerun score 0.80 and two trials from a harness swap 0.79, with no difference detected, while two trials from a model swap score 0.74, detectably less (Appendix~\ref{app:similarity}).} %
\rev{The harnesses give the model different tools, yet the same model mostly ends at the same files, and a different model less often does.}

\rev{We find no sign that a harness has a set of tasks it reliably wins.}
If one did, a task that only that harness passed in the first run should again be passed only by that harness in the second run.
\rev{We test this on the two Qwen models on hard45, which gives six pairs of harnesses with two runs on each side.}
Among the tasks on which a pair disagreed in both runs, \pn{rep.harness.same} went to the same harness both times and \pn{rep.harness.opp} went to the other harness the second time, about what chance alone would give.
For a model swap the split leans the other way: between Qwen3.8-27B and Qwen3.6-35B-A3B under the same harness, \pn{rep.model.same} of \pn{rep.model.both} such tasks went to the same model both times (Appendix~\ref{app:mde}).

The same flip rates set how small a difference each task set can detect.
Two runs on hard45 disagree on about 6 of the 45 tasks, and a paired test can draw only on those tasks. %
With so few of them, even a difference in which every one favoured the same harness would be only barely significant.
\rev{At this rate of disagreement, no difference on 45 tasks, however large, is detected four times in five, and a difference of \pn{mde.hard45.pp} points is detected only about half the time.}
The same calculation for the 445 tasks of pool447 gives a detectable difference of about \pn{mde.pool447.pp} points at 80\% probability (Figure~\ref{fig:mde} in Appendix~\ref{app:mde}).
Differences of a few tasks between harnesses on hard45 therefore cannot be told from zero one model at a time, which is why the bounds of the previous section come from pool447 and from averages over models.

\subsection{Harness effects exist, and their direction changes with the model}
\label{sec:losses}

\rev{Harness effects on pass rate exist in particular pairings of model and harness, and their direction changes with the model.}
\rev{The best-measured one is a loss: with Qwen3.8-27B on pool447, OpenCode trails Claude Code and mini-SWE-agent by 8 to 9 points, more than the \pn{mde.pool447.pp} points this task set can detect (Table~\ref{tab:pool447}).} %
\rev{A trial that ends at a limit is part of what a harness delivers, so this deficit is a harness effect on pass rate.} %
\rev{The trials point to the output cap, the limit on the length of one model response.}
\rev{Claude Code's responses were cut off at its cap in 52 trials on this task set, and it prompted the model to continue. OpenCode does not continue a response that the cap cuts off, and 34 of its trials ended that way, all but one of them failing, against one such trial for Claude Code and none for mini-SWE-agent (Appendix~\ref{app:cutoff}).} %

\rev{Two checks narrow the cause (Appendix~\ref{app:checks}).}
\rev{OpenCode had a shorter wall-clock limit than Claude Code on the Qwen models, and this does not explain the deficit: no OpenCode trial on pool447 reached the limit, and on hard45, where some did, rerunning OpenCode with Claude Code's limit removed every timeout and changed its passes by one trial.} %
\rev{Raising the output cap does help: on the 73 pool447 tasks where any OpenCode trial of the first round had ended on the cap, run twice with each cap, a larger cap lifts its passes from 86 to 99 of 146 trials ($p$ = 0.076), which is consistent with truncation losing tasks, short of significance.} %

\rev{The same harness leads on another model.}
\rev{With DeepSeek-V4.1-Flash on hard45, OpenCode passes 27 of the 45 tasks in each of two runs and mini-SWE-agent passes 21 and 20, a lead of 14 points (90\% interval 7 to 23).} %
\rev{On pool447 with the same model, the same two harnesses are under 2 points apart.} %
\rev{Across the seven models on hard45, the differences between OpenCode and each of the other two harnesses vary more than chance alone would give ($p$ = 0.013 and 0.019, uncorrected; Appendix~\ref{app:pairs}).}
\rev{A harness difference measured on one model, or on one task set, does not carry over to another.}

\rev{The other conditions that were not equal across harnesses, such as the compatibility layer through which Claude Code reaches other vendors' models, each have a check in Table~\ref{tab:conditions}.}

Each harness reaches its limits in its own way.
mini-SWE-agent has no context management, so a long trial overflows the model's context window: with Qwen3.8-27B on hard45 this ended \pn{out.qwen38.hard45.mini.overflow} of the 45 trials of a run, where Claude Code and OpenCode, which compact the history, lost at most one.
Most of those overflows fall on tasks that the other two harnesses failed as well, so they mark tasks the model was already losing (Figure~\ref{fig:outcomes} and Appendix~\ref{app:sensitivity}).

\subsection{Two benchmarks of other kinds give the same picture}

\rev{If the findings above belonged only to repository repair, a different kind of task should separate the harnesses.}
\rev{We ran DeepSeek-V4.1-Flash twice in every harness on SpreadsheetBench Verified and on the 150 DABstep tasks.} %
\rev{On SpreadsheetBench Verified the 90\% intervals exclude a difference beyond 2 points between any two harnesses.} %
\rev{On DABstep the order of the three harnesses changed between the two runs: Claude Code had the highest pass rate in the first run and the lowest in the second.} %
\rev{The pass rates are in Table~\ref{tab:results} and the paired differences in Appendix~\ref{app:bench}, and Appendix~\ref{app:tb} adds one run per harness on Terminal-Bench 2.1 with Qwen3.6-35B-A3B.}

\subsection{Cost differences between harnesses also depend on the model}

\begin{table}[t]
\centering
\caption{\captabcost}
\label{tab:cost}
\fitbox{\footnotesize\setlength{\tabcolsep}{4pt}
\begin{tabular}{@{}lccccc@{}}
\toprule
 & \multicolumn{3}{c}{Mean cost per trial} & \multicolumn{2}{c}{Highest $\div$ lowest} \\
\cmidrule(lr){2-4}\cmidrule(l){5-6}
Model, task set (currency) & Claude Code & mini-SWE-agent & OpenCode & per trial & per solved task \\
\midrule
GLM-5.3-Flash, hard45 (CNY) & 1.81 / 2.09 & 0.97 / 1.06 & 0.56 / 0.52 & 3.2$\times$ / 4.0$\times$ & 3.15$\times$ \\
HY4-Preview, hard45 (CNY) & 2.86 / 2.42 & 4.13 / 3.20 & 2.41 / 1.63 & 1.7$\times$ / 2.0$\times$ & 1.6$\times$ \\
DeepSeek-V4-Flash, hard45 (CNY) & 0.98 & 1.02 & 0.66 & 1.5$\times$ & 1.3$\times$ \\
DeepSeek-V4.1-Flash, hard45 (CNY) & 0.31 & 0.35 & 0.41 & 1.3$\times$ & 1.3$\times$ \\
DeepSeek-V4.1-Flash, pool447 (CNY) & 0.143 & 0.148 & 0.147 & 1.03$\times$ & 1.05$\times$ \\
Claude Opus 5.5, hard45 (USD) & 0.42 & 0.39 & 0.39 & 1.08$\times$ & 1.14$\times$ \\
\bottomrule
\end{tabular}
}
\end{table}

\rev{For the same model on the same tasks, the most expensive harness costs under 10\% more than the cheapest with DeepSeek-V4.1-Flash on pool447 and with Claude Opus 5.5 on hard45, and 3 to 4 times as much with GLM-5.3-Flash on hard45 (Table~\ref{tab:cost}).} %
\rev{With GLM-5.3-Flash and HY4-Preview, the second runs reproduce the first-run cost differences with the harnesses in the same order.} %
\rev{With GLM-5.3-Flash, where the difference is largest, the extra spending bought no detectable gain in passes: over two runs Claude Code passes 41 of 90 trials and OpenCode 36, and a solved task costs 4.28 CNY under Claude Code against 1.36 under OpenCode.} %

\begin{figure}[t]
\centering
\includegraphics[width=\linewidth]{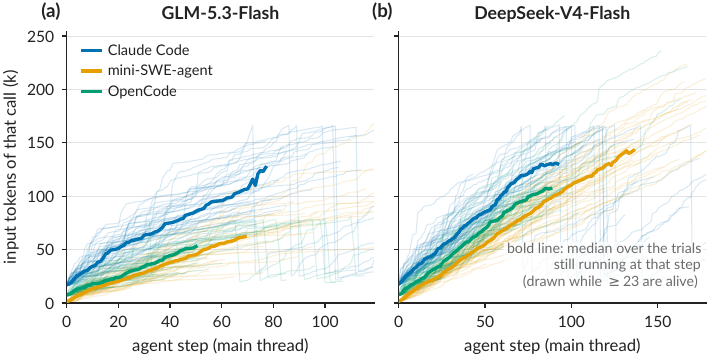}
\caption{\capfigctx}
\label{fig:ctx}
\end{figure}

\rev{The cost differences come from three terms: the preamble a harness sends with every step, what it keeps in the conversation history, and the number of steps the model takes.}
The preamble, which is the system prompt and the tool definitions, is 16.6k tokens in Claude Code, 7.0k in OpenCode, and 0.8k in mini-SWE-agent, and Claude Code also keeps the model's own reasoning in the history. %
\rev{Which term dominates depends on the model.}
With GLM-5.3-Flash on hard45, Claude Code and mini-SWE-agent take about as many steps, so their cost difference is the input per step.
With DeepSeek-V4-Flash on hard45, mini-SWE-agent takes \pn{cost.dsflash.mini.steps} steps to Claude Code's \pn{cost.dsflash.CC.steps}, its history grows with each one, and the lightest harness ends up consuming the most tokens (Figure~\ref{fig:ctx}, Appendix~\ref{app:budget}).

The price of cached input scales cost up or down without changing the order of the three harnesses.
Each step sends the history of the previous step again, so nearly all input tokens are cache hits in every harness, and the price of a hit decides how large a share of cost the history is.
The vendor of GLM-5.3-Flash charges \pn{price.glmflash.ratio} of the full input price for a cache hit and the vendor of DeepSeek-V4-Flash \pn{price.dsflash.ratio}, and pricing each model's tokens at the other vendor's prices keeps the order of the three harnesses in both cases (Appendix~\ref{app:budget}).
\rev{Appendix~\ref{app:equalbudget} rescores runs under an equal token budget by truncating finished trials after the fact, a post-hoc analysis.}

\subsection{Other checks}
\label{sec:checks}

Three further checks agree with the sections above.
DeepSeek's own command-line harness, run with DeepSeek-V4-Flash on hard45, passes \pn{dsflash.hard45.dsh.k} of the 45 tasks, inside the range of the three harnesses in Table~\ref{tab:results} (Appendix~\ref{app:dsh}).
Inside Claude Code, removing the file-edit tools, removing the shell, and using a release from a year earlier each leave Claude Opus 5 within one task of the unmodified runs on hard45 (Appendix~\ref{app:opus}).
Turning off thinking, the reasoning a model writes before it answers, with Qwen3.6-35B-A3B on hard45 costs mini-SWE-agent \pn{think.mini.drop.pp} points and moves Claude Code and OpenCode by \pn{think.cc.delta.pp} and \pn{think.oc.delta.pp} points, one more effect that depends on the pairing of model and harness (Appendix~\ref{app:thinking}).

\section{Discussion and Conclusion}

\rev{Which harness is better has no answer apart from the model.}
\rev{Across seven models no harness leads on average, and the harness with the largest deficit on one model has the largest lead on another, so a difference measured on one model does not carry over.}
\rev{Within the bounds our task sets give, a harness swap otherwise behaves like a rerun: it changes about as many outcomes and leads the model to the same files.}
\rev{Harness engineering shows up in two places: in gains and losses that belong to particular pairings, such as a cut-off response that one harness does not continue, and in cost, where the preamble, the history, and the number of steps weigh differently on each model.}
\rev{This is our answer to the title: what a harness changed most in our runs was the number of tokens a task consumed, up to 3 to 4 times in cost on one model, while outside particular pairings its effect on pass rate stayed within the bounds above.}

\rev{For evaluators, the consequence is to report a rerun with every harness difference and to name the model it was measured on.}
\rev{Differences of a few points read from single runs over a few dozen tasks lie within rerun variation, and on a larger task set a single run can still show a difference that the next run does not.}
A rerun of the same configuration is the cheapest control available, and it is the one most comparisons leave out.

\paragraph{Limitations.}
\rev{The two benchmarks outside repository repair were run with one model. On a single model, one run over 445 tasks does not reliably detect a difference below about 5 points, so smaller single-model effects remain open.} %

\paragraph{Conclusion.}
\rev{With the model held fixed, none of three production harnesses leads consistently, and a harness swap changes about as many task outcomes as a rerun.}
\rev{The exceptions are effects of particular pairings of model and harness, and they point in different directions on different models.}
\rev{Cost follows the same pattern: how much more one harness costs than another depends on the model.}
We recommend that every reported harness difference come with a rerun of the same configuration.

\section*{Ethics statement}
All tasks come from public benchmarks\rev{ (SWE-bench Verified, SpreadsheetBench Verified, DABstep, and Terminal-Bench 2.1)}, and no human subjects or personal data are involved.
The upstream fixes that networked agents retrieved are public GitHub content; we report retrieval rates by model to inform evaluation practice and make no claim that any vendor trained on or intended to use the test data.
API keys and credentials were removed from all released files.

\section*{Reproducibility statement}
Every number in the paper is generated from tables of trials by the scripts that draw the figures and tables, and Appendix~\ref{app:controls} lists the harness versions, the settings each configuration ran with, and how the offline condition was enforced.
The trial table, the analysis scripts, and the Harbor job files \rev{of the first round of runs}, with credentials removed, are available at \url{https://github.com/YangzeLiu/what-does-a-harness-buy}\rev{, and those of the second round will be added to the same repository}.
The full trajectories of the mini-SWE-agent and OpenCode runs, and the tool-call sequences and results of the Claude Code runs, will be released with the paper.

\section*{AI use statement}
In this work, we used generative AI tools to implement standard methods, to assist with data analysis, and to draft and edit parts of the paper. The agent trajectories this paper measures are model output by design and are the object of study itself. We have reviewed all AI-assisted work: the analysis scripts were rerun by the authors against the trial table behind every table and figure, and all drafted and edited text was read and approved by the authors. We take responsibility for the final content of this work, including text, claims or artifacts produced with the aid of generative AI.

\bibliography{refs}
\bibliographystyle{iclr2027_conference}

\appendix
\section*{Appendix}
\renewcommand{\thesection}{\Alph{section}}
\setcounter{section}{0}
\setcounter{table}{0}\renewcommand{\thetable}{A\arabic{table}}
\setcounter{figure}{0}\renewcommand{\thefigure}{A\arabic{figure}}

\FloatBarrier
\section{What was held fixed}
\label{app:controls}

Three settings were aligned across the harnesses on purpose, the step cap, the point at which the conversation is compacted, and network access, and every other setting was left as shipped.
Table~\ref{tab:controls} lists the configuration of each harness as run.

\begin{table}[htbp]
\centering
\caption{\textbf{What was held fixed.} Verified from the harness configurations, the Harbor job files, and the proxy logs.}
\label{tab:controls}
\footnotesize\setlength{\tabcolsep}{4pt}
\begin{tabular}{@{}>{\raggedright\arraybackslash}p{2.2cm}>{\raggedright\arraybackslash}p{3.9cm}>{\raggedright\arraybackslash}p{3.3cm}>{\raggedright\arraybackslash}p{3.7cm}@{}}
\toprule
 & Claude Code & mini-SWE-agent & OpenCode \\
\midrule
Version$^{a}$ & 2.1.259 to 2.1.273 (2.1.263 on GLM-5.3-Flash and Claude Opus~5\rev{, 2.1.280 on Claude Opus~5.5}) & 2.4.6 & 1.18.30 to 1.18.31 (1.18.30 on GLM-5.3-Flash) \\
Sampling & none sent (no flag) & none sent & none sent \\
Thinking / effort & GLM-5.3-Flash: effort \texttt{high} (command-line flag)\newline Qwen: effort \texttt{xhigh} (environment setting, ignored by vLLM)$^{b}$\newline DeepSeek: Claude Code's own default\rev{\newline Claude Opus~5.5: effort \texttt{high}} & GLM-5.3-Flash: \texttt{reasoning\_effort} = \texttt{high} (request body)\rev{\newline Claude Opus~5.5: \texttt{high}}\newline Qwen, DeepSeek: default & GLM-5.3-Flash: \texttt{reasoningEffort} = \texttt{high} (model configuration)\rev{\newline Claude Opus~5.5: \texttt{high}}\newline Qwen, DeepSeek: default \\
Tools & built-in set (Bash, Read, Edit, Write\ldots)\rev{, with no separate Grep tool while the shell is available}\newline \texttt{WebSearch} disallowed & \texttt{bash} only & bash, read, edit, write, glob, grep, webfetch, task\newline ripgrep pre-placed \\
Context management & auto-compaction: 110k window (environment setting) on the Qwen configurations, client default elsewhere$^{c}$ & none\newline per-observation head and tail trim (10k chars)\newline hard stop at 131k & prune (40k tool-output tail) and auto-compaction at the input limit minus a 20k reserve (78k)\newline input limit 98k, output limit 32k \\
Termination & \texttt{-{}-max-turns 300}\newline soft timeout (exit 124) & \texttt{step\_limit=300}\newline no soft timeout\rev{ (one added for DeepSeek-V4.1-Flash and Claude Opus~5.5)} & \texttt{build.steps=300}\newline soft timeout (exit 124) \\
Wall clock (safety net)$^{d}$ & Qwen: 12$\times$ the task's agent timeout\newline vendor API: \rev{soft timeout where set, otherwise a multiple of the task's agent timeout} & Qwen: 6$\times$ the task's agent timeout\newline vendor API: \rev{same} & \rev{Qwen: 6$\times$ the task's agent timeout, and a soft timeout of 5{,}760\,s on Qwen3.8-27B with hard45}\newline vendor API: \rev{same} \\
Network (agent phase) & allowlist: the model endpoint and the host machine, which serves a package mirror & same & same \\
Runner & Harbor 0.22.0, one container per trial, verifier after the agent exits & same & same \\
\bottomrule
\multicolumn{4}{@{}p{\linewidth}@{}}{\footnotesize $^{a}$~the versions of the configurations scored in Tables~\ref{tab:results} and \ref{tab:pool447}, read off every scored trial. OpenCode 1.18.25 to 1.18.29 appear only in OpenCode as shipped (Table~\ref{tab:sensitivity}) and in the ablations.  $^{b}$~a compatibility setting with no effect on the Qwen models: \texttt{xhigh} is the Qwen chat template's own default, and vLLM's Anthropic-compatible endpoint ignores the effort field (verified by a single-prompt probe against the same serving stack). Claude Code's reasoning per step is within 3 to 6\% of OpenCode's on the same model, and OpenCode sends no effort, so all three harnesses see the same thinking budget on the Qwen configurations.  $^{c}$~\texttt{CLAUDE\_CODE\_AUTO\_COMPACT\_WINDOW}${=}$110000 is set on the Qwen configurations only, where it fires at a measured median of 78k tokens and so matches OpenCode by construction. On the vendor-API configurations Claude Code keeps its own default window and compacts much later (on GLM-5.3-Flash at a median of \pn{budget.cc.compact.pre.k}k prompt tokens, Table~\ref{tab:budget}).  $^{d}$~\rev{on the locally served Qwen configurations Harbor's limit is twelve times the task's 3{,}000\,s agent timeout for Claude Code and six times for mini-SWE-agent and OpenCode, except that OpenCode on Qwen3.8-27B with hard45 ran under a soft timeout of 5{,}760\,s. On the vendor-API configurations the limit is a soft timeout where the run command carries one, 2{,}760\,s on the DeepSeek and Claude Opus models and 5{,}760\,s on GLM-5.3-Flash, and otherwise Harbor's limit, a multiple of the task's agent timeout.}} \\
\end{tabular}

\end{table}

\paragraph{Budget.}
\rev{The cap counts the steps of a harness's main conversation, and model calls made by a sub-agent come on top of it.}
\rev{The step cap is set through each harness's own knob:} \texttt{-{}-max-turns} in Claude Code, \texttt{step\_limit} in mini-SWE-agent, \texttt{build.steps} in OpenCode.
\rev{Wall-clock limits differ between the harnesses on the locally served Qwen configurations.}
\rev{There Harbor's limit is twelve times the task's 3,000\,s agent timeout for Claude Code, whose prefix-cache pattern against vLLM makes each step about twice as slow, and six times for mini-SWE-agent and OpenCode. The exception is OpenCode on Qwen3.8-27B with hard45, which ran under a soft timeout of 5,760\,s.}
\rev{On the vendor-API configurations the limit is a soft timeout where the run command carries one, 2,760\,s on the DeepSeek and Claude Opus models and 5,760\,s on GLM-5.3-Flash, and otherwise Harbor's limit, a multiple of the task's agent timeout.}
On Qwen3.8-27B with hard45, nine OpenCode trials of the first run ended at the 5,760\,s soft timeout, where Claude Code could run to 36,000\,s, and six of them failed. %
\rev{These trials count as OpenCode's result. When OpenCode is rerun with Claude Code's 36,000\,s, no trial ends at the wall clock, against 12 over the two original runs, and it passes 32 of 89 scored trials, against 31 of 89, so the wall clock does not explain its deficit (Table~\ref{tab:conditions}, Appendix~\ref{app:checks}). On pool447 no OpenCode trial was ended by the wall clock.} %
\rev{A trial that reaches a step or time limit is scored like any other, on the repository it leaves. The exception is a trial whose tests started while the agent was still running past the limit, which counts as lost to infrastructure.}

\paragraph{Context window.}
Claude Code and OpenCode compact the conversation at about the same size on the Qwen configurations, and mini-SWE-agent does not compact.
The two Qwen models are served with a 131k-token window.
OpenCode is configured with an input limit of 98k tokens and an output limit of 32k, so the two together fit the window, and it compacts when the conversation exceeds the input limit minus a 20k reserve, at about 78k tokens.
On the Qwen configurations Claude Code's auto-compaction window is set to 110k, which fires at a measured median of 78k tokens, matching OpenCode by construction.
On the vendor-API configurations Claude Code runs with its default window and compacts much later, on GLM-5.3-Flash at a median of 168k prompt tokens.
mini-SWE-agent has no compaction and runs into the model's window, and that outcome is counted in its own column of Table~\ref{tab:denominators}.

\paragraph{Network.}
The agent phase of every reported run reaches no network beyond the model endpoint and the host machine, which serves a package mirror, because web access lets agents retrieve the upstream fix (Appendix~\ref{app:leakage}).
On the servers this is Harbor's allowlist mode.
The Claude Opus runs go through a proxy that admits only the model API during the agent phase and any host during verification.
OpenCode downloads its search program, ripgrep, at first use, so its \texttt{grep} and \texttt{glob} tools fail offline. The reported OpenCode has the binary placed in advance, and OpenCode as shipped is kept for Table~\ref{tab:sensitivity}.

\paragraph{Other fixed conditions.}
OpenCode selects its system prompt by model-name pattern, and every model in this study falls into its default prompt.
Claude Code runs with a fresh configuration directory per trial and with web search disallowed.
\rev{Whenever the shell is available, Claude Code's tool list has no separate \texttt{Grep} tool for any model, Anthropic or not, so search runs through \texttt{Bash}.}
Thinking is on by default in every reported configuration, including the three HY4-Preview configurations, which are called through the vendor's API.
The Qwen3.6-35B-A3B thinking-off configurations of Appendix~\ref{app:thinking} switch it off at the serving layer, so the switch is the same for all three harnesses. The HY4-Preview thinking-off configuration switches it off through the API's thinking field, which the endpoint treats as on when it is missing, and it runs in Claude Code only.
For the HY4-Preview Claude Code configurations the reverse proxy folds Claude Code's mid-conversation system messages into the preceding user message and sets the thinking field explicitly, which aligns the vendor endpoint with the defaults of the Anthropic API. The content and the order of the conversation are unchanged.
\rev{Claude Code sends a reasoning-effort setting to every endpoint (\texttt{xhigh} on the Qwen configurations).} vLLM's Anthropic-compatible endpoint ignores it, and on the Qwen configurations the reasoning per step under Claude Code is within a few percent of OpenCode's, which sends none.

\FloatBarrier
\section{DeepSeek's own harness}
\label{app:dsh}

A vendor's own harness lands inside the range of the three general harnesses on pass rate and just above it on cost.
DeepSeek's command-line harness, release 0.1.2-rc.1, ran DeepSeek-V4-Flash on hard45, driven by Harbor like the other three.
All 45 trials were scored, and \pn{dsflash.hard45.dsh.k} passed, the same count as the first run of Claude Code on the same model. Six trials ended with a non-zero exit from the harness, three of them after a passing patch.
The harness takes \pn{steps.dsflash.hard45.dsh.mean} model calls per trial, between OpenCode's \pn{steps.dsflash.hard45.OC.mean} and mini-SWE-agent's \pn{steps.dsflash.hard45.mini.mean} on this model, and costs \pn{dsflash.hard45.dsh.cny} CNY per trial at DeepSeek's list price, 0.02 CNY above mini-SWE-agent, the most expensive of the three general harnesses on this model. %

\FloatBarrier
\section{Task sets and coverage}
\label{app:pools}

Every task set excludes the tasks whose verifier needs the network, and a trial lost to an infrastructure fault leaves the denominator.
Table~\ref{tab:denominators} traces the denominators from the 500 tasks of SWE-bench Verified to the two task sets, and counts, for the first run of each configuration in the first round, the trials that were lost to infrastructure and the trials that ended at a limit.
Table~\ref{tab:coverage} lists every configuration of the first round with its runs and its scored count.
\rev{Table~\ref{tab:results} in the main text gives every run of every configuration, with the runs of the second round in bold.}
The eight removed tasks are the \texttt{psf/requests} instances, whose verifier calls a public web service. hard45 and pool447 share no task.
Two scikit-learn tasks time out in the verifier in most configurations, which is why the pool447 configurations have 444 to 446 scored tasks.
Eight tasks in three configurations have two scored trials for the same run. The trial that finished last is used, and using the first instead changes no count.

\begin{table}[htbp]
\centering
\caption{\textbf{Denominators.} (a) From SWE-bench Verified to the two task sets. (b) For the first run of each configuration in the first round, the trials lost to infrastructure (removed from the denominator) and the trials that ended at a limit (kept and scored).}
\label{tab:denominators}
{\small\setlength{\tabcolsep}{3.5pt}
\begin{tabular}{@{}llrrrrrrrr@{}}
\toprule
\multicolumn{10}{@{}p{\linewidth}@{}}{\emph{(a) From SWE-bench Verified to the two task sets}} \\
\multicolumn{10}{@{}p{\linewidth}@{}}{SWE-bench Verified: \textbf{500} tasks} \\
\multicolumn{10}{@{}p{\linewidth}@{}}{\quad $-$ 8 \texttt{psf/requests} tasks, whose verifier calls a public web service and hangs offline $\rightarrow$ 492 usable tasks} \\
\multicolumn{10}{@{}p{\linewidth}@{}}{\quad $\rightarrow$ \textbf{hard45} $=$ 45 (the whole difficulty bands of 1 to 4\,h and over 4\,h, 42 $+$ 3) $+$ \textbf{pool447} $=$ 447 (the remaining tasks, disjoint from hard45)} \\
\addlinespace
\multicolumn{10}{@{}p{\linewidth}@{}}{\emph{(b) What the denominator loses, first run of each configuration in the first round}} \\
\addlinespace
 &  &  &  & not &  &  & step & output &  \\
Model & Harness & $n$ & scored & scored & overflow & timeout & cap & cap & other \\
\midrule
\multicolumn{10}{@{}l}{\emph{hard45}} \\
Qwen3.6-35B-A3B & Claude Code & 45 & 45 & 0 & 0 & 0 & 1 & 0 & 0 \\
 & mini-SWE-agent & 45 & 45 & 0 & 2 & 0 & 0 & 0 & 0 \\
 & OpenCode & 45 & 45 & 0 & 0 & 0 & 0 & 0 & 0 \\
Qwen3.8-27B & Claude Code & 45 & 43 & 2 & 1 & 0 & 0 & 0 & 0 \\
 & mini-SWE-agent & 45 & 45 & 0 & 11 & 0 & 0 & 0 & 0 \\
 & OpenCode & 45 & 45 & 0 & 0 & 6 & 0 & 0 & 0 \\
DeepSeek-V4-Flash & Claude Code & 45 & 45 & 0 & 0 & 0 & 0 & 0 & 0 \\
 & mini-SWE-agent & 45 & 45 & 0 & 0 & 0 & 0 & 0 & 0 \\
 & OpenCode & 45 & 45 & 0 & 0 & 0 & 0 & 0 & 1 \\
GLM-5.3-Flash & Claude Code & 45 & 45 & 0 & 0 & 1 & 0 & 0 & 0 \\
 & mini-SWE-agent & 45 & 45 & 0 & 0 & 0 & 0 & 0 & 0 \\
 & OpenCode & 45 & 45 & 0 & 0 & 0 & 0 & 0 & 0 \\
HY4-Preview & Claude Code & 45 & 45 & 0 & 0 & 0 & 0 & 0 & 0 \\
 & mini-SWE-agent & 45 & 45 & 0 & 0 & 0 & 0 & 0 & 0 \\
 & OpenCode & 45 & 45 & 0 & 0 & 0 & 0 & 0 & 0 \\
Claude Opus 5 & Claude Code & 45 & 45 & 0 & 0 & 0 & 0 & 0 & 0 \\
\addlinespace
\multicolumn{10}{@{}l}{\emph{pool447}} \\
Qwen3.6-35B-A3B & Claude Code & 447 & 445 & 2 & 0 & 0 & 2 & 0 & 0 \\
 & mini-SWE-agent & 447 & 445 & 2 & 6 & 0 & 0 & 0 & 0 \\
 & OpenCode & 447 & 446 & 1 & 0 & 0 & 0 & 0 & 0 \\
Qwen3.8-27B & Claude Code & 447 & 444 & 3 & 1 & 0 & 0 & 1 & 0 \\
 & mini-SWE-agent & 447 & 445 & 2 & 18 & 1 & 0 & 0 & 2 \\
 & OpenCode & 447 & 445 & 2 & 0 & 0 & 0 & 33 & 0 \\
\bottomrule
\multicolumn{10}{@{}p{\linewidth}@{}}{\footnotesize \emph{Not scored}: lost to an infrastructure fault and removed from the denominator. \emph{Overflow}: the conversation exceeded the model's context window. \emph{Timeout} and \emph{step cap}: the trial ended at the wall-clock or the step limit. \emph{Output cap}: the trial ended on the 32k output-token cap, for OpenCode with a final response cut at that size and for Claude Code with an error that ends the session. \emph{Other}: the harness exited with an error. \rev{The limit columns count the trials that ended at that limit and failed. A trial that ends at a limit is scored like any other, on the repository it leaves.} The wall clock differed between harnesses on the Qwen models, and Table~\ref{tab:conditions} and Appendix~\ref{app:checks} report the check.} \\
\end{tabular}
}

\end{table}

\begin{table}[htbp]
\centering
\caption{\textbf{Coverage of the first round of runs.} Every configuration, with its runs, the scored task count, and the passes of each run. The runs of the second round are in Table~\ref{tab:results}.}
\label{tab:coverage}
\begingroup\renewcommand{\small}{\scriptsize}{\small\setlength{\tabcolsep}{3.5pt}
\begin{tabular}{@{}lllcrrrr@{}}
\toprule
Model & Task set & Harness & Run & $n$ & scored & pass & not scored \\
\midrule
Qwen3.6-35B-A3B & hard45 & Claude Code & 1 & 45 & 45 & 11 & 0 \\
 &  & Claude Code & 2 & 45 & 45 & 10 & 0 \\
 &  & mini-SWE-agent & 1 & 45 & 45 & 15 & 0 \\
 &  & mini-SWE-agent & 2 & 45 & 45 & 13 & 0 \\
 &  & OpenCode & 1 & 45 & 45 & 11 & 0 \\
 &  & OpenCode & 2 & 45 & 45 & 11 & 0 \\
 & pool447 & Claude Code & 1 & 447 & 445 & 310 & 2 \\
 &  & mini-SWE-agent & 1 & 447 & 445 & 318 & 2 \\
 &  & mini-SWE-agent & 2 & 447 & 445 & 313 & 2 \\
 &  & OpenCode & 1 & 447 & 446 & 297 & 1 \\
\addlinespace
Qwen3.8-27B & hard45 & Claude Code & 1 & 45 & 43 & 19 & 2 \\
 &  & Claude Code & 2 & 45 & 45 & 20 & 0 \\
 &  & mini-SWE-agent & 1 & 45 & 45 & 17 & 0 \\
 &  & mini-SWE-agent & 2 & 45 & 45 & 18 & 0 \\
 &  & OpenCode & 1 & 45 & 45 & 17 & 0 \\
 &  & OpenCode & 2 & 45 & 44 & 14 & 1 \\
 & pool447 & Claude Code & 1 & 447 & 444 & 347 & 3 \\
 &  & mini-SWE-agent & 1 & 447 & 445 & 353 & 2 \\
 &  & OpenCode & 1 & 447 & 445 & 312 & 2 \\
\addlinespace
DeepSeek-V4-Flash & hard45 & Claude Code & 1 & 45 & 45 & 21 & 0 \\
 &  & Claude Code & 2 & 45 & 45 & 23 & 0 \\
 &  & mini-SWE-agent & 1 & 45 & 45 & 22 & 0 \\
 &  & OpenCode & 1 & 45 & 45 & 19 & 0 \\
\addlinespace
GLM-5.3-Flash & hard45 & Claude Code & 1 & 45 & 45 & 22 & 0 \\
 &  & mini-SWE-agent & 1 & 45 & 45 & 20 & 0 \\
 &  & OpenCode & 1 & 45 & 45 & 18 & 0 \\
\addlinespace
HY4-Preview & hard45 & Claude Code & 1 & 45 & 45 & 23 & 0 \\
 &  & mini-SWE-agent & 1 & 45 & 45 & 26 & 0 \\
 &  & OpenCode & 1 & 45 & 45 & 21 & 0 \\
\addlinespace
Claude Opus 5 & hard45 & Claude Code & 1 & 45 & 45 & 36 & 0 \\
 &  & Claude Code & 2 & 45 & 45 & 35 & 0 \\
 &  & Claude Code & 3 & 45 & 45 & 35 & 0 \\
\bottomrule
\multicolumn{8}{@{}p{\linewidth}@{}}{\footnotesize First round of runs only. The runs of the second round are in Table~\ref{tab:results}, in bold. $n$ is the size of the task set, \emph{scored} counts the trials that were scored, and \emph{not scored} counts the trials lost to an infrastructure fault, which leave the denominator and the paired analyses. The reported OpenCode has ripgrep placed in advance (Appendix~\ref{app:sensitivity}). Configurations outside this grid, namely thinking off, the other Claude Code settings on Claude Opus 5, DeepSeek's command-line harness, OpenCode as shipped, and every web-enabled run, are reported in their own sections.} \\
\end{tabular}
}
\endgroup
\end{table}

\FloatBarrier
\section{Leakage audit of the web-enabled runs}
\label{app:leakage}

Agents with network access look up the upstream fix, so every pass rate in the paper comes from runs without it.
Before the offline condition was enforced, the DeepSeek and Qwen configurations ran with network access.
An audit of those runs scanned tool-call arguments for retrieval of the upstream fix: a GitHub search hit, a commit patch, a pull-request diff, or the post-fix file from the raw content host.
Of 439 trials, 312 had fetched the fix, 152 of them citing the task's own issue or pull-request number, and 310 of the 312 went on to edit a file after the lookup.
The Claude models did this in 4 of 445 trials.
Pass rate followed: 96\% of the scored DeepSeek and Qwen trials that retrieved the fix passed, against 71\% of those that never touched the network.
All web-enabled pass-rate results were discarded, and the condition was moved from the prompt to the network layer.

\FloatBarrier
\section{Per-task outcomes}
\label{app:heatmap}

\begin{figure}[htbp]
\centering
\includegraphics[width=0.85\linewidth]{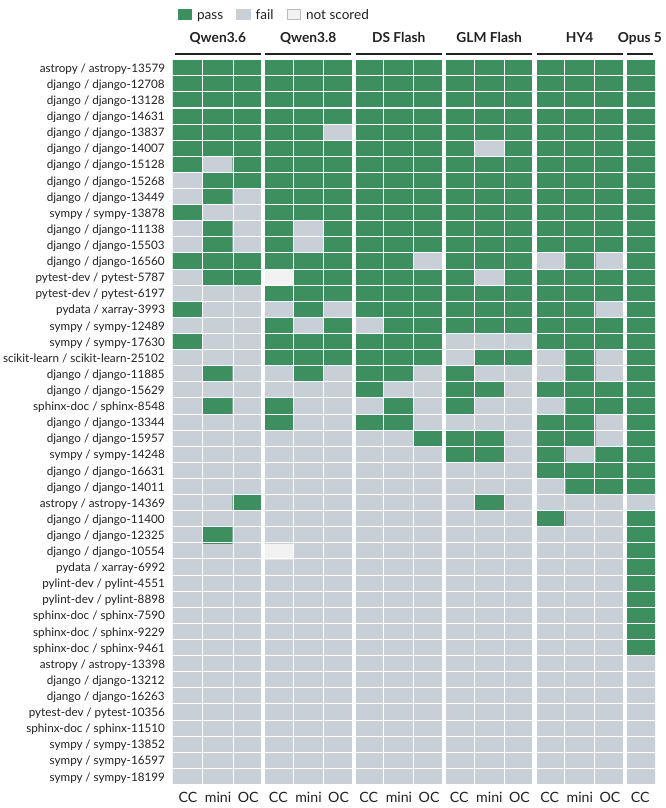}
\caption{\textbf{hard45, task by task.} First run of every configuration in the first round. \rev{CC, mini and OC are Claude Code, mini-SWE-agent and OpenCode, and DS Flash and GLM Flash are DeepSeek-V4-Flash and GLM-5.3-Flash.} Eight tasks are passed by no configuration and four by every configuration.}
\label{fig:heatmap}
\end{figure}

Figure~\ref{fig:heatmap} shows every hard45 task under every model and harness.
Rows are sorted by how many configurations passed the task.

\FloatBarrier
\section{How small a difference a paired comparison can detect}
\label{app:mde}

A paired comparison of two harnesses draws only on the tasks on which their runs disagree, and on hard45 there are too few of them to detect any difference with 80\% power.

\paragraph{Flip rates.}
The flip rate of a pair of runs is the share of tasks whose outcome differs between them, $(b+c)/n$, where $b$ counts the tasks passed only in the first run, $c$ those passed only in the second, and $n$ the tasks scored in both.
In the first round the median flip rate on hard45 is \pn{flip.hard45.rerun.median}\% across the \pn{flip.hard45.rerun.n} rerun pairs, and \pn{flip.hard45.rerun.ladder.median}\% across the \pn{flip.hard45.rerun.ladder.n} pairs that leave out Claude Opus~5, whose three runs agree most closely.
It is \pn{flip.hard45.harness.median}\% across the \pn{flip.hard45.harness.n} harness swaps, and \pn{flip.hard45.model.median}\% at the median and up to \pn{flip.hard45.model.max}\% across the model swaps.
\rev{With the second round included (Figure~\ref{fig:noise}) the medians are 13.3\% across 22 rerun pairs, 13.3\% across 21 harness swaps and 25.6\% across 70 model swaps. Without the pairs of the two Claude Opus models they are 15.6\% across 16 rerun pairs and 13.3\% across 18 harness swaps.}
On pool447 the one rerun pair of the first round flips \pn{flip.pool447.rerun.median}\% of tasks and the six harness swaps \pn{flip.pool447.harness.min} to \pn{flip.pool447.harness.max}\%.
\rev{On DeepSeek-V4.1-Flash, run twice in every harness on pool447, a rerun flips 6.3 to 7.6\% of tasks and a harness swap 6.9 to 7.8\%.} %

\paragraph{Replication of task-level wins.}
A task that one harness wins in the first run goes to that harness again in the second run about as often as chance would give.
The replication test of the main text starts from the \pn{rep.harness.split1} disagreements in run 1 of the six harness pairs with two runs on each side, on \pn{rep.harness.tasks} distinct tasks.
Among the \pn{rep.harness.both} tasks on which a pair disagrees in both runs, \pn{rep.harness.same} go to the same harness both times and \pn{rep.harness.opp} to the other one, a binomial $p$ = \pn{rep.harness.p}, and $p$ = \pn{rep.harness.pclust} under a permutation that flips the sign of a whole task at once.
\rev{The same test applied to a model swap, Qwen3.8-27B against Qwen3.6-35B-A3B under the same harness, gives \pn{rep.model.same} tasks that go to the same model both times and \pn{rep.model.opp} that go to the other, among \pn{rep.model.both} ($p$ = \pn{rep.model.p}).}

\begin{figure}[htbp]
\centering
\includegraphics[width=0.6\linewidth]{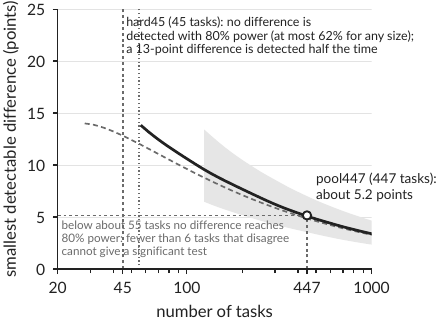}
\caption{\textbf{Smallest paired difference detected with 80\% power, against the number of tasks.} Exact McNemar test at $\alpha$ = .05, at the \pn{mde.hard45.disc}\% rate of disagreement pooled over the harness swaps of the first round (solid line) and across the range observed between harness pairs (band). The asymptotic formula is dashed. At 45 tasks no difference is detected with 80\% power.}
\label{fig:mde}
\end{figure}

\paragraph{Power.}
The smallest paired difference the exact McNemar test detects at 80\% power depends on $n$ and on the rate at which two runs disagree. Figure~\ref{fig:mde} plots it at the \pn{mde.hard45.disc}\% rate pooled over the harness swaps of the first round, with a band for the range across pairs.
Power is unconditional: the number of tasks on which the two runs disagree is binomial in $n$ and that rate, and given that number the two-sided exact test at $\alpha$ = .05 rejects or not.
Fewer than six such tasks pointing the same way cannot reject at all, so below \pn{mde.hard45.nmin80} tasks no difference reaches 80\% power at this rate. At 45 tasks the highest power for any difference is \pn{mde.hard45.powermax}\%, reached when every such task points the same way, and a difference of \pn{mde.hard45.pp50} points is detected half the time.
On the 447 tasks of pool447 the difference detected with 80\% power is \pn{mde.pool447.ppexact} points.
The asymptotic formula that inverts the usual sample-size expression is drawn dashed for reference and is optimistic by half a point to a point everywhere.

\paragraph{Equivalence margins.}
\rev{The equivalence margin was first set at $\pm$10 points and was tightened to $\pm$5 after pool447 had run, to the difference that task set detects with 80\% power, about \pn{mde.pool447.ppexact} points.}
\rev{Table~\ref{tab:pool447} reports, for each difference, the narrowest of $\pm$2, $\pm$3 and $\pm$5 points that contains its whole 90\% interval.}
The Claude Code minus mini-SWE-agent interval is [\pn{qwen36.pool447.CCvmini.lonine}, \pn{qwen36.pool447.CCvmini.hinine}] on Qwen3.6-35B-A3B and [\pn{qwen38.pool447.CCvmini.lonine}, \pn{qwen38.pool447.CCvmini.hinine}] on Qwen3.8-27B, both inside $\pm$5.
\rev{Two runs of DeepSeek-V4.1-Flash disagree on fewer tasks, and at its rate of disagreement the 445 scored tasks of pool447 detect a difference of 3.7 points with 80\% power.} %

\FloatBarrier
\section{Where the per-step input goes}
\label{app:budget}

Claude Code's larger input per step comes from what it carries into every call, its preamble and the model text it keeps in the history, since a single tool return is about the same size in all three harnesses.
Table~\ref{tab:budget} rebuilds the input each harness sends per step on GLM-5.3-Flash, the one model where all three harnesses ran the same 45 tasks with the same settings and complete trajectories.
Three notes on how it was measured.
The fixed cost per call is the prompt size of the first call, because the system prompts and tool schemas of Claude Code and OpenCode are not in the trajectories. The harness preamble is that size minus the issue statement, which is the same text in every harness, with mini-SWE-agent's instance template counted as preamble.
Claude Code's tool returns are taken from its raw stream log, because the trajectory converter appends the SDK's tool-result metadata to every observation and about doubles its size.
The reconstruction uses each harness's own empirical characters-per-token ratio, and for Claude Code it reproduces the measured input only if the model's reasoning blocks are counted as re-sent history. That term is inferred from the fit and is not observed, since the provider's usage does not separate reasoning from text.
The reconstructed input per step lands within \pn{budget.cc.perstep.gap}, \pn{budget.mini.perstep.gap}, and \pn{budget.oc.perstep.gap} of the measured input for the three harnesses.

The per-step growth behind Figure~\ref{fig:ctx} is the median over trials of the per-trial slope of input tokens against step index: on GLM-5.3-Flash \pn{ctx.glmflash.CC.slope} tokens per step under Claude Code against \pn{ctx.glmflash.mini.slope} under mini-SWE-agent and \pn{ctx.glmflash.OC.slope} under OpenCode, and on DeepSeek-V4-Flash \pn{ctx.dsflash.CC.slope} against \pn{ctx.dsflash.mini.slope} and \pn{ctx.dsflash.OC.slope}. The reasoning Claude Code keeps in the transcript is \pn{ctx.glmflash.CC.reason.step} and \pn{ctx.dsflash.CC.reason.step} tokens of that per step on the two models.
Repricing each model's tokens under the other vendor's list keeps the order of the harnesses: on the GLM tokens Claude Code costs \pn{reprice.glmflash.zhipu.CCvmini}$\times$ mini-SWE-agent at Zhipu's prices and \pn{reprice.glmflash.ds.CCvmini}$\times$ at DeepSeek's, and on the DeepSeek tokens \pn{reprice.dsflash.ds.CCvmini}$\times$ and \pn{reprice.dsflash.zhipu.CCvmini}$\times$. The DeepSeek-V4-Flash trial costs \pn{cost.dsflash.CC.cny} CNY under Claude Code and \pn{cost.dsflash.mini.cny} under mini-SWE-agent at DeepSeek's list price.

\begin{figure}[htbp]
\centering
\includegraphics[width=\linewidth]{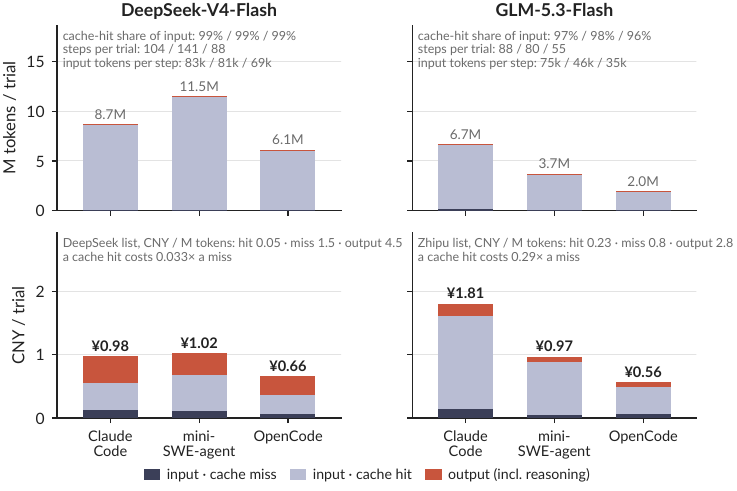}
\caption{\textbf{Where the cost comes from.} Top: tokens per trial on DeepSeek-V4-Flash and GLM-5.3-Flash, divided into cache-miss input, cache-hit input, and output including reasoning. Bottom: the same tokens priced at the vendor's list price. Annotations give the cache-hit share, steps per trial, and input tokens per step for the three harnesses.}
\label{fig:cost}
\end{figure}

\begin{table}[htbp]
\centering
\caption{\textbf{Where the per-step input goes.} GLM-5.3-Flash on hard45, 45 trials per harness. (a) Size of a single tool return. (b) The per-step input, rebuilt from the trajectories and compared with the provider's measured usage.}
\label{tab:budget}
{\footnotesize\setlength{\tabcolsep}{4pt}
\begin{minipage}{\linewidth}
\begin{tabular}{@{}lrrrrl@{}}
\multicolumn{6}{@{}l}{\emph{(a) Tool-return size per call (characters)}} \\
\toprule
Harness & calls & mean & median & 90th pct. & Largest sources \\
\midrule
Claude Code & 4{,}403 & 1{,}237 & 392 & 3{,}095 & \texttt{Bash} 68\%, \texttt{Read} 30\% \\
mini-SWE-agent & 3{,}787 & 1{,}186 & 461 & 3{,}121 & \texttt{bash} 100\% \\
OpenCode & 2{,}633 & 1{,}103 & 311 & 3{,}049 & \texttt{read} 49\%, \texttt{bash} 39\%, \texttt{grep} 9\% \\
\bottomrule
\end{tabular}

\vspace{5pt}

\begin{tabular}{@{}lrrr@{}}
\multicolumn{4}{@{}l}{\emph{(b) Per-step input budget (tokens, means over the 45 hard45 trials)}} \\
\toprule
 & Claude Code & mini-SWE-agent & OpenCode \\
\midrule
Fixed cost per call$^{a}$ & 17{,}134 & 1{,}382 & 7{,}579 \\
\quad of which the issue statement & 554 & 554 & 554 \\
\quad of which harness preamble & 16{,}581 & 829 & 7{,}025 \\
\addlinespace
Model text $+$ reasoning per trial$^{b}$ & 58{,}359 & 1{,}715 & 21{,}900 \\
Tool returns (observations) per trial$^{b,c}$ & 31{,}557 & 44{,}871 & 16{,}781 \\
Tool-call arguments per trial$^{b}$ & 11{,}719 & 14{,}554 & 5{,}130 \\
\quad characters per token (empirical)$^{b}$ & 3.84 & 2.22 & 3.85 \\
\addlinespace
Steps per trial & 88.1 & 80.2 & 55.1 \\
Trials with an auto-compaction (of 45)$^{d}$ & 8 & 0 & 0 \\
\addlinespace
Reconstructed input per step & 80{,}060 & 39{,}933 & 33{,}068 \\
\textbf{Measured input per step}$^{e}$ & \textbf{75{,}286} & \textbf{45{,}897} & \textbf{35{,}039} \\
\bottomrule
\multicolumn{4}{@{}p{\linewidth}@{}}{\footnotesize GLM-5.3-Flash on hard45, 45 trials per harness. The model, its settings and the task set are identical across the three columns, so every difference here belongs to the harness.  $^{a}$~the first call's prompt tokens, which every later call also pays before any transcript: the harness's system prompt and tool schemas, its own user-message wrapper, and the issue statement. The preamble is what remains after subtracting the issue statement at 3.7~characters per token, measured on the first user turn, which OpenCode and Claude Code receive verbatim and identically task by task (2{,}049~characters, 554~tokens on average). mini-SWE-agent's user turn also carries its instance template (a constant 2{,}860~characters), which is counted as harness preamble. The harnesses' own system prompts are in neither the trajectory nor the agent log, so the preamble is a residual.  $^{b}$~characters appended to the conversation over one trial, converted at the harness's own empirical rate, fitted on the step-to-step change in prompt tokens against the change in characters. Claude Code and OpenCode carry prose at about 3.8~characters per token, and mini-SWE-agent's raw shell output is token-dense at 2.2. Everything a step appends is re-sent with every later request. The fit closes for Claude Code only if its reasoning blocks are re-sent too, which we infer from the fit and do not observe directly.  $^{c}$~Claude Code's observation sizes are read from its raw stream log, because the trajectory converter appends the SDK's tool-result metadata to every observation and roughly doubles it, and that metadata was never sent to the model. mini-SWE-agent and OpenCode trajectories match their raw logs and are used as stored.  $^{d}$~auto-compaction boundaries recorded by the Claude Code SDK. mini-SWE-agent and OpenCode write no such record and never reset in these trials, so Claude Code's reconstruction is an upper bound and the other two are exact. On this model Claude Code runs with its own default auto-compaction window (the 110k setting of Table~\ref{tab:controls} applies to the Qwen configurations only) and compacts at a median of 168k prompt tokens, where OpenCode compacts at its input limit minus a 20k reserve, at 78k, which is why the compaction counts differ.  $^{e}$~input tokens reported by the provider, averaged over all steps of all 45 trials.} \\
\end{tabular}
\end{minipage}
}

\end{table}

\FloatBarrier
\section{The output cap and a resume prompt for OpenCode}
\label{app:cutoff}

OpenCode loses the trials that its output cap cuts off, because it does not prompt the model to continue, while Claude Code prompts the model to continue and almost never ends a trial on the cap.
OpenCode's provider configuration caps a single response at 32k output tokens, and the trajectories show the cut landing at exactly that size in the middle of a reasoning block.
On pool447 with Qwen3.8-27B, 38 of 447 OpenCode trials contain such a cut and in 34 of them it is the final step, all but one of which fail. Those trials pass at 7.9\% against 75.9\% for the rest, and 32 of the 34 end with the repository untouched, after a median of 11 steps.
Claude Code on the same model and task set is cut in 52 of 447 trials, 66 events in all. Of these, 64 are followed by the harness's own resume prompt and the session continues, and one trial ends on the cap.
Its trials with a cut pass at 69\% against 79\% without.
mini-SWE-agent never reaches the cap, and its largest single step is 13k output tokens.

Giving OpenCode Claude Code's resume prompt does not detectably recover these tasks.
We modified Harbor's OpenCode wrapper to re-invoke the session with Claude Code's resume string, verbatim, whenever the last step was cut or the repository was untouched. We then reran the 41 pool447 tasks whose first trial had failed that way under both the unmodified and the modified wrapper, two trials each, interleaved on one host.
The mean per-task pass rate is 0.585 unmodified and 0.634 with the resume prompt. Six tasks improve, two worsen and 33 tie ($p$ = .29, exact sign test).
The prompt fired in 15\% of the modified trials, against the 33 of 41 first trials that had been cut, so whether a trial is cut varies from run to run.
Recovery cost 1.23$\times$ the steps and 1.30$\times$ the input tokens.
\rev{Raising the cap itself from 32k to 131k output tokens on the affected tasks is the second check of Appendix~\ref{app:checks}.}

\FloatBarrier
\section{Claude Code settings on Claude Opus 5}
\label{app:opus}

Claude Opus~5 passes within one task of its control runs in every Claude Code setting we tried, and the year-old release changes only the cost.
Claude Code with Claude Opus~5 was run on hard45 in four settings (Table~\ref{tab:opus}): three control runs of the pinned release, the file-editing tools removed so that the model edits through the shell, run twice, the shell removed, and release 1.0.100, a year older than the pinned one.
Every setting passes between \pn{opus.noshell.k} and \pn{opus.bashonly.k2} of 45, within one task of the control runs' \pn{opus.ctrl.k2} to \pn{opus.ctrl.k1}.
Against the three control runs, \pn{opus.bashonly1.flips} and \pn{opus.bashonly2.flips} tasks flip without the editing tools, \pn{opus.noshell.flips} without the shell, and \pn{opus.v1100.flips} with the old release, against \pn{opus.ctrl.floor} between the control runs themselves.
The old release does not enable extended thinking by default, so its median output per trial is \pn{opus.v1100.outratio}$\times$ that of the controls and its cost per trial \pn{opus.v1100.costratio}$\times$.

\begin{table}[htbp]
\centering
\caption{\textbf{Claude Code on Claude Opus~5, four settings, hard45.} Flips are counted against the three control runs. Cost is what Claude Code itself reports, at Anthropic's list price.}
\label{tab:opus}
\fitbox{{\footnotesize\setlength{\tabcolsep}{3.5pt}
\begin{tabular}{@{}llrcrrrrrr@{}}
\toprule
 & Claude Code & \multicolumn{2}{c}{Passed / 45} & flips & McNemar & output & reasoning & steps & USD per trial \\
Setting & version & $k$ & 95\% interval (\%) & of 135 & $p$ vs.\ run 1 & tokens & tokens & & (mean) \\
\midrule
Control, run 1 & 2.1.263 & 36 & [66.2, 89.1] & n/a & n/a & 15{,}984 & 4{,}916 & 34 & 1.83 \\
Control, run 2 & 2.1.263 & 35 & [63.7, 87.5] & n/a & 1.00 & 12{,}777 & 5{,}927 & 32 & 1.51 \\
Control, run 3 & 2.1.263 & 35 & [63.7, 87.5] & n/a & 1.00 & 14{,}930 & 5{,}345 & 39 & 1.60 \\
\addlinespace
Bash only, run 1 & 2.1.263 & 36 & [66.2, 89.1] & 16 & 1.00 & 13{,}539 & 5{,}378 & 31 & 1.42 \\
Bash only, run 2 & 2.1.263 & 37 & [68.7, 90.7] & 7 & 1.00 & 12{,}250 & 5{,}853 & 28 & 1.63 \\
No shell & 2.1.263 & 34 & [61.3, 85.8] & 4 & 0.50 & 16{,}734 & 10{,}031 & 26 & 1.86 \\
Claude Code 1.0.100 & 1.0.100 & 34 & [61.3, 85.8] & 12 & 0.62 & 3{,}351 & 0 & 26 & 0.73 \\
\bottomrule
\multicolumn{10}{@{}p{\linewidth}@{}}{\footnotesize Claude Opus~5 through Claude Code on hard45, 45 tasks per row, no web access. Every row is one run, and all 45 tasks are scored in every row.  \emph{Control} is the configuration of the main tables, run three times.  \emph{Bash only} removes the file-editing tools through Claude Code's disallowed-tools setting (\texttt{Edit}, \texttt{Write}, \texttt{MultiEdit}, \texttt{NotebookEdit} and \texttt{WebSearch}), so the model can edit only through the shell. It was run twice.  \emph{No shell} removes execution instead (\texttt{WebSearch} and \texttt{Bash} disallowed) and leaves the editing tools.  \emph{Claude Code 1.0.100} is the release of 2025-09-01, one year older than the 2.1.263 of the other six rows, with every other setting identical.  \emph{Passed} counts the tasks passed, with a Wilson 95\% interval.  \emph{Flips} counts the tasks whose outcome differs from a control run, summed over the three control runs ($3\times45$). Between the three control runs themselves the same count is 6 of 135, or 4.4\%.  \emph{McNemar} is the two-sided exact binomial test against control run~1 on the tasks on which the two runs disagree.  \emph{Output} and \emph{reasoning tokens} are per-trial medians of the provider's output-token and reasoning-token counts. Release 1.0.100 predates extended thinking being on by default, which is where its 4$\times$ smaller output and its zero reasoning come from.  \emph{USD per trial} is the mean of the cost Claude Code reports for itself, at Anthropic's list price for the tokens used.} \\
\end{tabular}
}
}
\end{table}

\FloatBarrier
\section{Thinking on and off}
\label{app:thinking}

Switching thinking off hurts Qwen3.6-35B-A3B under mini-SWE-agent and leaves it about where it was under Claude Code and OpenCode, so what thinking is worth depends on the harness.
Thinking was switched off at the serving layer, with a chat template that closes the thinking block at once and a ban on the opening token, so the switch is the same for all three harnesses.
Each harness ran hard45 twice with thinking on and twice with it off, and Table~\ref{tab:thinking} pairs the per-task mean pass over the two runs.
Claude Code and OpenCode move by \pn{think.cc.delta.pp} and \pn{think.oc.delta.pp} points, each with \pn{think.cc.up} tasks going up and \pn{think.cc.down} down.
mini-SWE-agent moves by \pn{think.mini.delta.pp} points, with \pn{think.mini.down} tasks going down and \pn{think.mini.up} up.
Without thinking, Claude Code and OpenCode take more steps, \pn{think.cc.steps.on}$\to$\pn{think.cc.steps.off} and \pn{think.oc.steps.on}$\to$\pn{think.oc.steps.off} at the median, and mini-SWE-agent fewer.

On HY4-Preview the same switch was thrown inside Claude Code alone, through the API's thinking field, with one run per side.
The model passes \pn{think.hy4.on.k} of 45 with thinking on and \pn{think.hy4.off.k} with it off, \pn{think.hy4.on.pct}$\to$\pn{think.hy4.off.pct}\%, a drop of \pn{think.hy4.drop.pp} points. \pn{think.hy4.down} tasks are lost and \pn{think.hy4.up} gained ($p$ = \pn{think.hy4.p}).
Output falls by more than half, \pn{think.hy4.outk.on}k to \pn{think.hy4.outk.off}k tokens per trial, and the median step count rises, \pn{think.hy4.steps.on}$\to$\pn{think.hy4.steps.off}.
The cost follows the tokens, \pn{think.hy4.cny.on} to \pn{think.hy4.cny.off} CNY per trial at the vendor's list price.

\begin{table}[htbp]
\centering
\caption{\textbf{Thinking on and off, hard45.} Top block: Qwen3.6-35B-A3B, three harnesses, two runs per side, thinking switched off at the serving layer. The per-task mean is paired over the tasks both sides scored, and $p$ is a two-sided sign test on the tasks whose mean changed. The OpenCode configurations here run without ripgrep in place. Bottom block: HY4-Preview, Claude Code only, one run per side, thinking switched off through the API's thinking field. With one run per side the same sign test is McNemar's exact test on the \pn{think.hy4.down} lost and \pn{think.hy4.up} gained tasks.}
\label{tab:thinking}
{\footnotesize\setlength{\tabcolsep}{3.5pt}
\begin{tabular}{@{}lrrcrrrcc@{}}
\toprule
Harness & \multicolumn{2}{c}{passed / trials} & per-task mean & $\Delta$ & up / & sign & median steps & median output \\
 & on & off & on \,$\to$\, off & (points) & down & $p$ & on \,$\to$\, off & on \,$\to$\, off \\
\midrule
Claude Code & 21/90 & 22/87 & 0.233\,$\to$\,0.256 & +2.2 & 7\,/\,5 & 0.77 & 103.5\,$\to$\,126 & 27{,}761\,$\to$\,24{,}550 \\
mini-SWE-agent & 28/90 & 18/90 & 0.311\,$\to$\,0.200 & $-$11.1 & 2\,/\,10 & 0.04 & 106.5\,$\to$\,98 & 25{,}198\,$\to$\,20{,}582 \\
OpenCode & 15/88 & 16/86 & 0.148\,$\to$\,0.182 & +3.4 & 7\,/\,5 & 0.77 & 56\,$\to$\,78.5 & 14{,}018\,$\to$\,15{,}949.5 \\
\addlinespace
\multicolumn{9}{@{}l@{}}{\emph{HY4-Preview, Claude Code only, one run per side, thinking switched at the API}} \\
Claude Code & 23/45 & 18/45 & 0.511\,$\to$\,0.400 & $-$11.1 & 4\,/\,9 & 0.27 & 75\,$\to$\,90 & 43{,}189\,$\to$\,15{,}001 \\
\bottomrule
\multicolumn{9}{@{}p{\linewidth}@{}}{\footnotesize Qwen3.6-35B-A3B on hard45, no web access, two independent runs per side and per harness (90 trials). Thinking is switched off at the server, so all three harnesses see the same model: a chat template closes the thinking block before generation, no reasoning parser runs, and the opening thinking token is banned during sampling.  \emph{Passed} counts the trials that passed over the trials that were scored. Seven thinking-off and two thinking-on trials were lost to verifier faults and were not rerun, which is why some denominators are below 90. One task was lost in both thinking-off runs of OpenCode, so the OpenCode pairing is over 44 tasks.  \emph{Per-task mean} is the mean outcome of a task over the scored trials of that side, and the comparison is paired over the tasks both sides scored ($n$ $=$ 45, 45 and 44). $\Delta$ is off minus on, in points.  \emph{Up / down} counts the tasks whose per-task mean rose or fell. The sign test is the two-sided exact binomial on those tasks, and the remaining tasks are ties. This is one comparison of two runs per side on 45 tasks.  Both OpenCode configurations here run without ripgrep in place, because the thinking-off configuration had none either, so they differ from the OpenCode of the main tables.  Claude Code carries reasoning effort \texttt{xhigh}, its setting on the Qwen configurations.  mini-SWE-agent records a copy of its visible message text as reasoning, so its reasoning length is not comparable with the other two and is omitted.  The second block is a different experiment: HY4-Preview through the vendor's API under Claude Code, one run per side on the same 45 tasks, with thinking switched off by setting the API's thinking field to disabled (the endpoint treats a missing field as on).  With one run per side a per-task mean is 0 or 1, so \emph{up / down} counts the tasks on which the two runs disagree, and the sign test on them is McNemar's exact test.  At the vendor's list price the trial costs 2.86 CNY with thinking on and 2.14 with it off, from mean output of 43.9k and 17.5k tokens per trial.} \\
\end{tabular}
}

\end{table}

\FloatBarrier
\section{Sensitivity of the paired differences}
\label{app:sensitivity}

\begin{figure}[htbp]
\centering
\includegraphics[width=\linewidth]{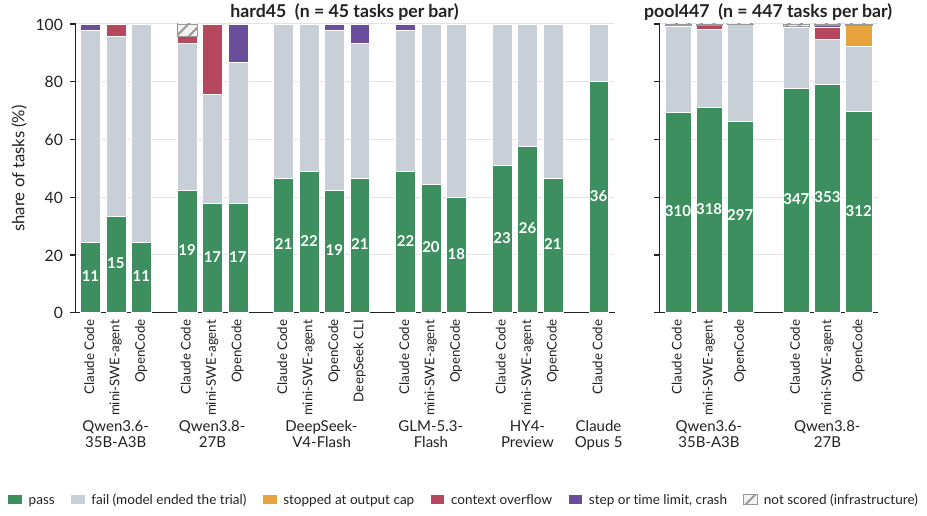}
\caption{\textbf{Most failed trials end with the model stopping on its own.} How each trial of the first run in the first round ended, as a share of tasks, with the number of passed tasks printed in each bar. The largest exceptions are mini-SWE-agent on Qwen3.8-27B, whose trials overflow the context window, and OpenCode on Qwen3.8-27B, whose trials stop at the output cap on pool447 and at the step or time limit on hard45. Not scored marks trials lost to infrastructure. DeepSeek CLI is DeepSeek's command-line harness (Appendix~\ref{app:dsh}).}
\label{fig:outcomes}
\end{figure}

No conclusion of the paper changes under four other ways of computing the paired differences.
Table~\ref{tab:sensitivity} recomputes every paired difference of the first round after dropping the trials that overflowed the context window, after scoring per protocol (dropping the tasks on which either configuration ended at a limit or was lost), with OpenCode as shipped in place of the reported OpenCode, and with the second run in place of the first wherever one exists.
\rev{The reported OpenCode has its search program, ripgrep, placed in the container in advance, because OpenCode downloads it at first use and the download fails offline. OpenCode as shipped is the sensitivity check of the third column.}
On pool447 the only entries that move by more than 3 points are the two OpenCode differences on Qwen3.8-27B under per-protocol scoring, which is the output-cap effect of Appendix~\ref{app:cutoff}.
On hard45 eleven entries move by more than 3 points, all smaller than what 45 tasks can detect (Appendix~\ref{app:mde}). The largest is OpenCode as shipped on Qwen3.6-35B-A3B, 11 points below the reported OpenCode, and one sign reverses, Claude Code minus mini-SWE-agent on DeepSeek-V4-Flash with the second run, by one task each way.

mini-SWE-agent's context overflows on Qwen3.8-27B fall mostly on tasks the model was already losing (Figure~\ref{fig:outcomes}).
Of the \pn{ovf.qwen38.hard45.mini.fail} hard45 tasks it lost that way, \pn{ovf.qwen38.hard45.mini.elsewhere} were passed by either other harness, and of the \pn{ovf.qwen38.pool447.mini.fail} it lost on pool447, \pn{ovf.qwen38.pool447.mini.elsewhere}. Recovering every one of them would move it by \pn{ovf.qwen38.hard45.mini.elsewhere.pp} and \pn{ovf.qwen38.pool447.mini.elsewhere.pp} points, and in its second run \pn{ovf.qwen38.hard45.mini.run2.again} of the \pn{ovf.qwen38.hard45.mini.run2.present} overflow again.

\begin{table}[htbp]
\centering
\caption{\textbf{Sensitivity of every paired difference.} Baseline: the first run of each configuration in the first round (Tables~\ref{tab:results} and~\ref{tab:pool447}). The other columns recompute the difference after dropping overflowed trials, scoring per protocol, using OpenCode as shipped, and using the second run. $=$ means the variant removes no task and leaves the value unchanged.}
\label{tab:sensitivity}
{\footnotesize\setlength{\tabcolsep}{3pt}
\begin{tabular}{@{}lrrrrr@{}}
\toprule
 & \multicolumn{1}{c}{Baseline difference} & No & Per & OpenCode & \\
Pair & \multicolumn{1}{c}{(points) [95\% interval] ($n$)} & overflow & protocol & as shipped & Run 2 \\
\midrule
\multicolumn{6}{@{}l}{\emph{hard45}} \\
\multicolumn{6}{@{}l}{Qwen3.6-35B-A3B} \\
\quad Claude Code $-$ mini-SWE-agent & $-$8.9 {\scriptsize [$-$23.9, +6.6] (45)} & $-$9.3$^{\scriptscriptstyle 43}$ & $-$6.8$^{\scriptscriptstyle 44}$ & n/a & $-$6.7 \\
\quad Claude Code $-$ OpenCode & +0.0 {\scriptsize [$-$11.8, +11.8] (45)} & $=$ & +0.0$^{\scriptscriptstyle 44}$ & +11.1 & $-$2.2 \\
\quad mini-SWE-agent $-$ OpenCode & +8.9 {\scriptsize [$-$4.0, +21.5] (45)} & +9.3$^{\scriptscriptstyle 43}$ & $=$ & +20.0 & +4.4 \\
\addlinespace
\multicolumn{6}{@{}l}{Qwen3.8-27B} \\
\quad Claude Code $-$ mini-SWE-agent & +7.0 {\scriptsize [$-$5.5, +19.1] (43)} & +0.0$^{\scriptscriptstyle 32}$ & +7.0 & n/a & +4.4$^{\scriptscriptstyle 45}$ \\
\quad Claude Code $-$ OpenCode & +7.0 {\scriptsize [$-$1.8, +15.5] (43)} & +7.1$^{\scriptscriptstyle 42}$ & +8.1$^{\scriptscriptstyle 37}$ & +4.7 & +11.4$^{\scriptscriptstyle 44}$ \\
\quad mini-SWE-agent $-$ OpenCode & +0.0 {\scriptsize [$-$11.2, +11.2] (45)} & +5.9$^{\scriptscriptstyle 34}$ & +0.0$^{\scriptscriptstyle 39}$ & $-$2.2 & +6.8$^{\scriptscriptstyle 44}$ \\
\addlinespace
\multicolumn{6}{@{}l}{DeepSeek-V4-Flash} \\
\quad Claude Code $-$ mini-SWE-agent & $-$2.2 {\scriptsize [$-$10.5, +6.2] (45)} & $=$ & $=$ & n/a & +2.2$^{\ast}$ \\
\quad Claude Code $-$ OpenCode & +4.4 {\scriptsize [$-$6.7, +15.3] (45)} & $=$ & +4.5$^{\scriptscriptstyle 44}$ & +0.0 & +8.9$^{\ast}$ \\
\quad mini-SWE-agent $-$ OpenCode & +6.7 {\scriptsize [$-$3.6, +16.6] (45)} & $=$ & +6.8$^{\scriptscriptstyle 44}$ & +2.2 & n/a \\
\addlinespace
\multicolumn{6}{@{}l}{GLM-5.3-Flash} \\
\quad Claude Code $-$ mini-SWE-agent & +4.4 {\scriptsize [$-$6.7, +15.3] (45)} & $=$ & +6.8$^{\scriptscriptstyle 44}$ & n/a & n/a \\
\quad Claude Code $-$ OpenCode & +8.9 {\scriptsize [$-$2.2, +19.5] (45)} & $=$ & +11.4$^{\scriptscriptstyle 44}$ & n/a & n/a \\
\quad mini-SWE-agent $-$ OpenCode & +4.4 {\scriptsize [$-$6.7, +15.4] (45)} & $=$ & $=$ & n/a & n/a \\
\addlinespace
\multicolumn{6}{@{}l}{HY4-Preview} \\
\quad Claude Code $-$ mini-SWE-agent & $-$6.7 {\scriptsize [$-$18.2, +5.2] (45)} & $=$ & $=$ & n/a & n/a \\
\quad Claude Code $-$ OpenCode & +4.4 {\scriptsize [$-$6.7, +15.3] (45)} & $=$ & $=$ & n/a & n/a \\
\quad mini-SWE-agent $-$ OpenCode & +11.1 {\scriptsize [$-$0.7, +22.2] (45)} & $=$ & $=$ & n/a & n/a \\
\addlinespace
\multicolumn{6}{@{}l}{\emph{pool447}} \\
\multicolumn{6}{@{}l}{Qwen3.6-35B-A3B} \\
\quad Claude Code $-$ mini-SWE-agent & $-$1.8 {\scriptsize [$-$5.2, +1.6] (445)} & $-$1.8$^{\scriptscriptstyle 439}$ & $-$1.8$^{\scriptscriptstyle 443}$ & n/a & $-$0.7$^{\ast}$ \\
\quad Claude Code $-$ OpenCode & +2.9 {\scriptsize [$-$0.9, +6.7] (445)} & $=$ & +3.2$^{\scriptscriptstyle 443}$ & +5.0$^{\scriptscriptstyle 421}$ & n/a \\
\quad mini-SWE-agent $-$ OpenCode & +4.7 {\scriptsize [+1.1, +8.4] (445)} & +5.0$^{\scriptscriptstyle 439}$ & +4.7 & +6.7$^{\scriptscriptstyle 421}$ & +3.6$^{\ast}$ \\
\addlinespace
\multicolumn{6}{@{}l}{Qwen3.8-27B} \\
\quad Claude Code $-$ mini-SWE-agent & $-$1.4 {\scriptsize [$-$4.2, +1.5] (444)} & $-$2.4$^{\scriptscriptstyle 425}$ & $-$1.4$^{\scriptscriptstyle 440}$ & n/a & n/a \\
\quad Claude Code $-$ OpenCode & +7.9 {\scriptsize [+4.3, +11.5] (444)} & +7.9$^{\scriptscriptstyle 443}$ & +3.4$^{\scriptscriptstyle 410}$ & +5.9 & n/a \\
\quad mini-SWE-agent $-$ OpenCode & +9.2 {\scriptsize [+5.5, +12.9] (445)} & +10.5$^{\scriptscriptstyle 427}$ & +4.6$^{\scriptscriptstyle 409}$ & +7.0 & n/a \\
\bottomrule
\multicolumn{6}{@{}p{\linewidth}@{}}{\footnotesize Each entry is the paired difference (first harness minus second) over the tasks both configurations scored. The baseline is the first run of each configuration in the first round (Table~\ref{tab:results}, and Table~\ref{tab:pool447} on pool447, here with a 95\% interval). It keeps every scored trial, including those that overflowed the context window, reached a limit or had their output cut.  \emph{No overflow}: tasks on which either configuration exceeded the context window are dropped.  \emph{Per protocol}: tasks on which either configuration ended at a step cap, a timeout or an output cut, or lost its trial to an infrastructure fault, are dropped.  $=$ marks a variant that is defined but drops no task, and so reproduces the baseline exactly.  \emph{OpenCode as shipped}: OpenCode without ripgrep in place replaces the reported OpenCode.  \emph{Run 2}: the second run replaces the first.  n/a means the variant does not exist for that pair (no such trial, no as-shipped OpenCode, or no second run).  $^{\ast}$~only one of the two configurations has a second run, and the other stays at its first run.  A superscript is the variant's $n$ where it differs from the baseline's. Table~\ref{tab:pairs} averages over all runs instead.} \\
\end{tabular}
}

\end{table}

\FloatBarrier
\section{All paired differences on hard45}
\label{app:pairs}

\rev{Averaged over seven models, no pair of harnesses differs detectably on hard45, and the differences of single models point both ways.}
\rev{Table~\ref{tab:pairs} gives the paired difference for every model and every pair of harnesses. For each model, a task's outcome in a configuration is first averaged over that configuration's runs, and the difference is then taken task by task. The intervals come from a bootstrap that resamples tasks 10,000 times, and the average row weights the seven models equally.} %
\rev{All three averages contain zero, and their 95\% intervals exclude an average difference above 5 points. Single models range from $-$14.4 to +8.9 points.} %

\rev{OpenCode's standing against the other two harnesses changes with the model.}
\rev{The last row of the table tests whether the models disagree, by Cochran's $Q$ with standard errors from the bootstrap. The models share the same tasks, so its $p$ is nominal.} %
\rev{For the two pairs with OpenCode the models disagree by more than sampling error would produce ($Q$ = 16.1 and 15.1 on 6 degrees of freedom, $p$ = 0.013 and 0.019), and for Claude Code against mini-SWE-agent they do not ($Q$ = 10.6, $p$ = 0.10).} %

\begin{table}[htbp]
\centering
\caption{\textbf{Paired differences on hard45, every model and every run.} Difference in points, first harness minus second, with a 95\% interval from a bootstrap over tasks. A task's outcome is averaged over the runs of each configuration before the difference is taken.}
\label{tab:pairs}
\fitbox{\footnotesize\setlength{\tabcolsep}{4pt}
\begin{tabular}{@{}l rc rc rc@{}}
\toprule
 & \multicolumn{2}{c}{Claude Code $-$ mini-SWE-agent} & \multicolumn{2}{c}{Claude Code $-$ OpenCode} & \multicolumn{2}{c}{mini-SWE-agent $-$ OpenCode} \\
\cmidrule(lr){2-3}\cmidrule(lr){4-5}\cmidrule(l){6-7}
Model & diff. & 95\% interval & diff. & 95\% interval & diff. & 95\% interval \\
\midrule
Qwen3.6-35B-A3B & $-$7.8 & [$-$16.7, +1.1] & $-$1.1 & [$-$10.0, +7.8] & +6.7 & [$-$1.1, +14.4] \\
Qwen3.8-27B & +5.6 & [$-$1.1, +12.2] & +8.9 & [+2.2, +16.7] & +3.3 & [$-$2.2, +10.0] \\
DeepSeek-V4-Flash & +0.0 & [$-$5.6, +5.6] & +6.7 & [$-$3.3, +16.7] & +6.7 & [$-$2.2, +15.6] \\
GLM-5.3-Flash & +3.3 & [$-$2.2, +8.9] & +5.6 & [$-$1.1, +13.3] & +2.2 & [$-$4.4, +8.9] \\
HY4-Preview & $-$3.3 & [$-$8.9, +2.2] & +2.2 & [$-$3.3, +8.9] & +5.6 & [+0.0, +12.2] \\
DeepSeek-V4.1-Flash & +6.7 & [$-$2.2, +16.7] & $-$7.8 & [$-$16.7, +1.1] & $-$14.4 & [$-$25.6, $-$4.4] \\
Claude Opus 5.5 & $-$1.5 & [$-$10.4, +7.4] & $-$3.7 & [$-$8.1, $-$0.0] & $-$2.2 & [$-$11.1, +4.4] \\
\midrule
Average of the seven & +0.4 & [$-$2.1, +2.9] & +1.5 & [$-$1.5, +4.7] & +1.1 & [$-$1.4, +3.7] \\
Do the models disagree? ($p$) & \multicolumn{2}{c}{0.10} & \multicolumn{2}{c}{0.013} & \multicolumn{2}{c}{0.019} \\
\bottomrule
\end{tabular}
}
\end{table}

\FloatBarrier
\section{Second round of runs}
\label{app:newmodels}

\rev{The second round leaves no harness ahead across models: the largest difference of the first round does not recur, a new model favours OpenCode on hard45, and a frontier model run through its native interface gives Claude Code at most a small lead.}

\paragraph{Second runs of GLM-5.3-Flash and HY4-Preview.}
\rev{Both models ran a second time in all three harnesses on hard45, with the configurations of the first run. The largest difference between harnesses in the first round, five tasks on HY4-Preview, did not recur, and in the second run all three harnesses pass 24. On GLM-5.3-Flash the second run passes 19, 18 and 18 under Claude Code, mini-SWE-agent and OpenCode, against 22, 20 and 18 in the first. Across every configuration with two runs on hard45, a rerun moves the pass count by at most three tasks.} %

\paragraph{DeepSeek-V4.1-Flash.}
\rev{The vendor retired DeepSeek-V4-Flash, and its API name now serves DeepSeek-V4.1-Flash, which we treat as a separate model. It ran twice in every harness on hard45 and on pool447. The three harnesses share one soft timeout of 2,760\,s and keep their own default output cap, and no trial reached the timeout or a step limit.} %
\rev{On hard45 the two runs pass 24 and 23 tasks under Claude Code, 21 and 20 under mini-SWE-agent, and 27 and 27 under OpenCode.} %
\rev{On pool447, 445 of the 447 tasks are scored in every run. Claude Code passes 368 and 374 tasks in its two runs, mini-SWE-agent 368 and 360, and OpenCode 369 and 375.} %
\rev{A single run can mislead. mini-SWE-agent and OpenCode differ by 0.2 points in the first run and by 3.4 points in the second (McNemar $p$ = 0.008, or 0.024 after Holm correction over the three pairs). Pooled over both runs the difference is 1.8 points, with $p$ = 0.16 after Holm correction.} %

\begin{figure}[h]
\centering
\includegraphics[width=0.9\linewidth]{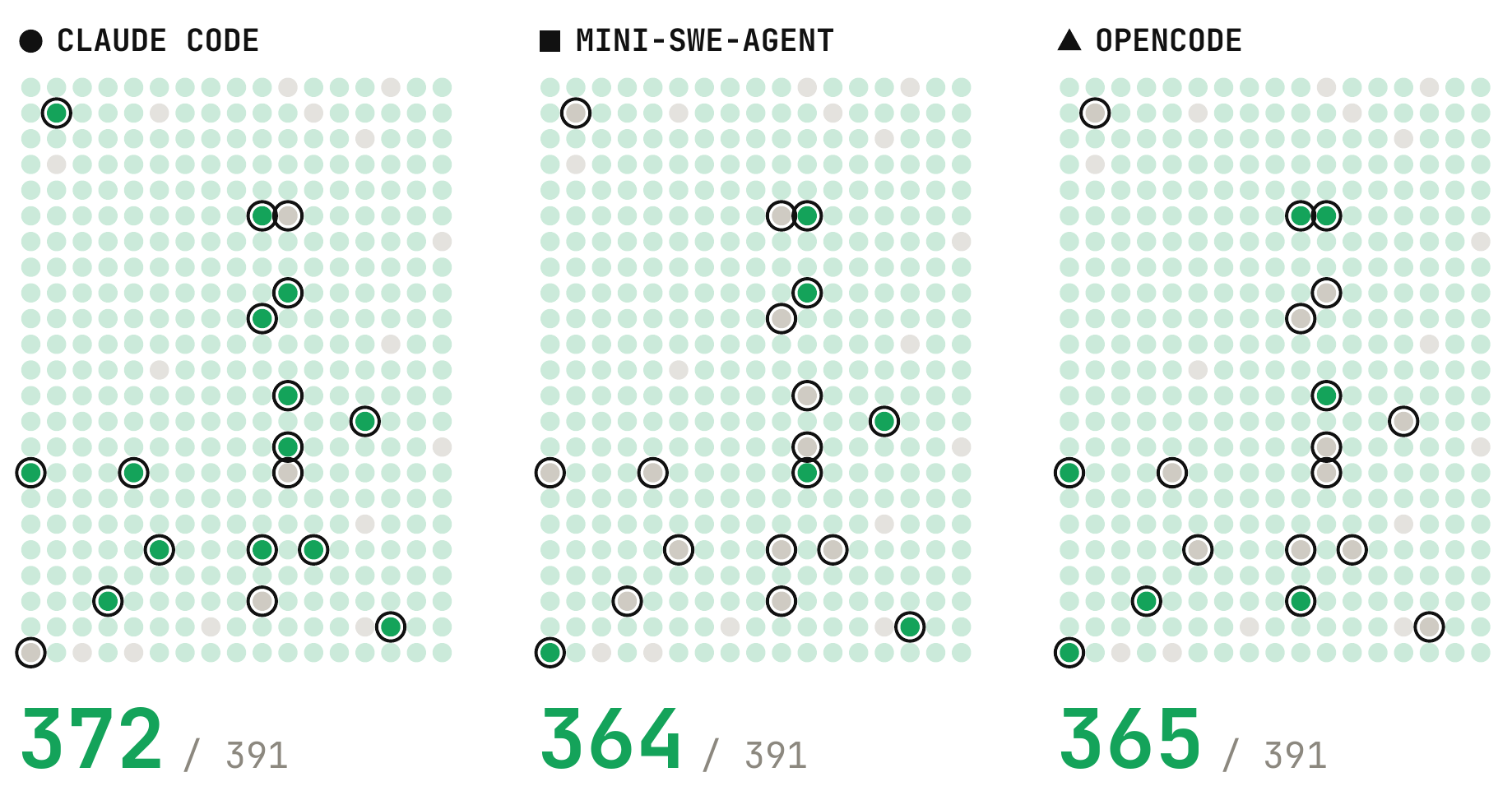}
\caption{\textbf{Claude Opus~5.5 on pool447, task by task.} Each dot is one of the 391 tasks that all three harnesses completed, at the same position in every panel. Green is a pass and grey a fail. On 373 tasks the three harnesses have the same outcome (faded dots): 358 are passed by all three and 15 by none. The 18 tasks on which they differ are ringed, and the differences between the pass counts under the panels come from these tasks alone.}
\label{fig:opus391}
\end{figure}

\paragraph{Claude Opus 5.5 through the native interface.}
\rev{Claude Opus~5.5 ran in all three harnesses through Anthropic's native interface, so no compatibility layer sits between Claude Code and the model. Claude Code ran at release 2.1.280. On hard45 Claude Code passes 36, 37 and 36 tasks in three runs, mini-SWE-agent 37 and OpenCode 38.} %
\rev{On pool447 the tasks ran in a pre-drawn random order, and the comparison uses the 391 tasks that all three harnesses completed. Claude Code completed all 447 tasks and passed 422.} %
\rev{On the 391 tasks the three harnesses pass 372, 364 and 365. Claude Code passes 11 tasks that mini-SWE-agent fails and fails 3 that it passes, a difference of +2.0 points (90\% interval [+0.4, +3.9], McNemar $p$ = 0.057, or 0.17 after Holm correction). Against OpenCode the counts are 10 and 3, a difference of +1.8 points ([+0.2, +3.5], $p$ = 0.092, or 0.18 after Holm correction). mini-SWE-agent and OpenCode differ on 4 and 5 tasks, $-$0.3 points ([$-$1.7, +1.2]). Figure~\ref{fig:opus391} shows the comparison task by task.} %
\rev{Through the native interface, Claude Code's lead is at most 4 points on pool447 and absent on hard45.}

\FloatBarrier
\section{Two checks on OpenCode's deficit on Qwen3.8-27B}
\label{app:checks}

\rev{The shorter wall clock does not explain OpenCode's deficit on Qwen3.8-27B, a larger output cap recovers part of it without a significant result, and the same harness leads on another model.}
\rev{Table~\ref{tab:conditions} lists every condition that was not equal across the harnesses, with its check.}

\begin{table}[htbp]
\centering
\caption{\captabconditions}
\label{tab:conditions}
\footnotesize\setlength{\tabcolsep}{4pt}
\begin{tabular}{@{}>{\raggedright\arraybackslash}p{0.29\linewidth}>{\raggedright\arraybackslash}p{0.30\linewidth}>{\raggedright\arraybackslash}p{0.35\linewidth}@{}}
\toprule
Condition that differed & Check & Result \\
\midrule
With Qwen3.8-27B on hard45 OpenCode had 5{,}760\,s of wall clock and Claude Code 36{,}000\,s &
OpenCode rerun at 36{,}000\,s (Qwen3.8-27B, hard45, two runs) &
Trials ended by the wall clock fall from 12 to 0; passes go from 31/89 to 32/89. On pool447 no OpenCode trial was ended by the wall clock. It does not explain the deficit. \\
\addlinespace
OpenCode does not continue a response cut off by its output cap &
Cap of 32k against 131k output tokens on the 73 affected pool447 tasks (Qwen3.8-27B, two runs each) &
Passes go from 86/146 to 99/146 ($p = 0.076$): consistent with truncation losing tasks, short of significance. \\
\addlinespace
For non-Anthropic models Claude Code runs through a compatibility layer &
Claude Opus 5.5 through the native interface in all three harnesses &
hard45: Claude Code 36, 37, 36; mini-SWE-agent 37; OpenCode 38. pool447 (391 tasks): 95.1\%, 93.1\%, 93.4\%. \\
\addlinespace
Timeout and output cap (rows 1 and 2) &
DeepSeek-V4.1-Flash with one soft timeout for all three harnesses and the same 32k output cap in Claude Code and OpenCode (pool447, two runs) &
83.4\%, 81.8\%, 83.6\%: within 2 points under equal settings. \\
\bottomrule
\end{tabular}

\end{table}

\paragraph{Equal wall clock.}
\rev{OpenCode reran hard45 twice with Claude Code's 36,000\,s of wall clock, 90 trials. Trials ended by the wall clock fall from 12 to none, and passes are 32 of 89 scored trials against 31 of 89 in the original runs, with one trial per side that could not be scored. Had the shorter wall clock caused the deficit, passes would have risen.} %

\paragraph{Larger output cap.}
\rev{On the 73 pool447 tasks on which a trial of any OpenCode configuration of the first round, OpenCode as shipped included, had ended on its cap, we alternated a cap of 32k and one of 131k output tokens, two runs each, with 36,000\,s of wall clock. Passes are 86 of 146 with the 32k cap and 99 of 146 with 131k. Per task, the larger cap passes more often on 18 tasks and less often on 8, with 47 ties (two-sided sign test, $p$ = 0.076). The result is consistent with truncation losing tasks and falls short of significance.} %
\rev{We ran this check on OpenCode only, because in the first round 34 of its trials on this task set ended on the cap, 33 of them failing, against one for Claude Code and none for mini-SWE-agent.} %

\paragraph{The same harness leads elsewhere.}
\rev{On DeepSeek-V4.1-Flash, where the three harnesses share one soft timeout, OpenCode leads mini-SWE-agent by 14.4 points on hard45 over two runs and by 1.8 points on pool447 (Table~\ref{tab:pairs}, Appendix~\ref{app:newmodels}). OpenCode's deficit on Qwen3.8-27B is a harness effect whose direction depends on the model.} %

\FloatBarrier
\section{Do the harnesses edit the same files?}
\label{app:similarity}

\rev{After a harness swap the agent edits about the same source files as after a rerun, and after a model swap it edits detectably different ones.}
\rev{Five models have reruns in all three harnesses on hard45: Qwen3.6-35B-A3B, Qwen3.8-27B, GLM-5.3-Flash, HY4-Preview and DeepSeek-V4.1-Flash. For two trials on the same task we take the sets of source files that each one has modified when it ends and compute their Jaccard similarity, where 1 means identical sets. A rerun pairs two runs of one configuration, a harness swap two harnesses under one model, and a model swap two models under one harness. Within each task the five models weigh equally, and the 95\% intervals come from 4,000 bootstrap resamples of tasks.} %
\rev{Two trials from a rerun score 0.796, two from a harness swap 0.786 and two from a model swap 0.735. A harness swap minus a rerun is $-$0.010 ([$-$0.025, +0.004]), so no difference is detected, and a model swap minus a rerun is $-$0.061 ([$-$0.088, $-$0.037]), detectably less. Model by model, a harness swap minus a rerun ranges from $-$0.027 to +0.001, and every one of the five intervals contains zero.} %
\rev{The differences are small in absolute terms because most hard45 fixes touch one file and most trials edit that file.} %

\FloatBarrier
\section{Cost per solved task and equal budgets}
\label{app:equalbudget}

\rev{Cost per solved task follows cost per trial, because the harnesses pass at similar rates, and an equal token budget orders the harnesses by how many tokens they spend.}

\paragraph{Cost per solved task.}
\rev{The largest difference is on GLM-5.3-Flash, where over two runs a solved task costs 4.28 CNY under Claude Code (95\% interval [3.03, 6.43]) and 1.36 CNY under OpenCode ([0.92, 2.19]), a ratio of 3.15 with intervals that do not overlap. On DeepSeek-V4.1-Flash over the 445 scored tasks of pool447 the three harnesses are within 5\% of one another per solved task (Table~\ref{tab:cost}).} %

\begin{figure}[htbp]
\centering
\includegraphics[width=0.55\linewidth]{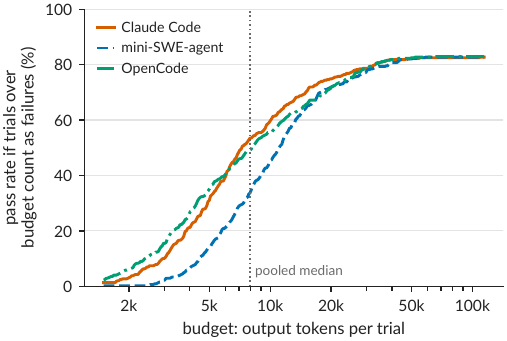}
\caption{\textbf{Pass rate under an equal output-token budget, by post-hoc truncation.} DeepSeek-V4.1-Flash, first run on the 445 scored tasks of pool447. A trial counts as passed at a budget if it passed and used at most that many output tokens. The dotted line marks the median output of the three harnesses together. No trial was rerun under a cap.}
\label{fig:equalbudget}
\end{figure}

\paragraph{Equal budgets by post-hoc truncation.}
\rev{To compare the harnesses at an equal budget we rescored the existing trials after the fact, counting a trial as failed when its output exceeds a common number of tokens. This is a post-hoc truncation of finished trials, and no trial was rerun under a cap.} %
\rev{On the first run of DeepSeek-V4.1-Flash on pool447 the three harnesses pass within 0.2 points of one another without a budget (82.7\%, 82.7\% and 82.9\%). With the budget at the median output of the three harnesses together, about 7.9k tokens, Claude Code passes 53.5\%, OpenCode 49.2\% and mini-SWE-agent 33.7\%, the order of their median output per trial, 6.4k, 7.3k and 9.9k tokens (Figure~\ref{fig:equalbudget}).} %
\rev{Once a budget binds, the order follows each harness's output per trial, so an equal budget measures how much a harness spends.}

\FloatBarrier
\section{Two benchmarks of other kinds}
\label{app:bench}

\rev{On two benchmarks outside software engineering no difference between the harnesses is detected either.}
\rev{SpreadsheetBench Verified has 400 real spreadsheet tasks, scored by comparing the workbook the agent produces with the reference. DABstep asks data-analysis questions over a set of financial data files, and we drew 150 of its hard questions at random. Both are scored deterministically. DeepSeek-V4.1-Flash ran each benchmark twice in each harness, and every trial was scored.} %
\rev{On SpreadsheetBench Verified the two runs pass 353 and 354 tasks under Claude Code, 355 and 348 under mini-SWE-agent, and 352 and 355 under OpenCode. Pooled over both runs, Claude Code minus mini-SWE-agent is +0.5 points ([$-$0.9, +1.9]), Claude Code minus OpenCode 0.0 ([$-$1.2, +1.2]) and mini-SWE-agent minus OpenCode $-$0.5 ([$-$1.9, +0.9]), so every 90\% interval excludes a difference above 2 points.} %
\rev{On DABstep the two runs pass 111 and 112 of 150 under Claude Code, 108 and 114 under mini-SWE-agent, and 107 and 113 under OpenCode. The pooled differences for the same three pairs are +0.3, +1.0 and +0.7 points, with 90\% intervals [$-$3.8, +4.4], [$-$3.4, +5.4] and [$-$3.5, +4.8], and the order of the harnesses changes between the runs: Claude Code is first in the first run and last in the second.} %
\rev{As on SWE-bench Verified, a harness swap flips about as many tasks as a rerun, 4.6 to 5.4\% against 3.8 to 5.8\% on SpreadsheetBench Verified, and 17.0 to 18.7\% against 14.0 to 17.3\% on DABstep.} %

\FloatBarrier
\section{A transfer check on Terminal-Bench 2.1}
\label{app:tb}

On Terminal-Bench 2.1, as on hard45, no ordering of the three harnesses is detectable.
We ran the three harnesses once with Qwen3.6-35B-A3B on the 53 Terminal-Bench 2.1 tasks whose reference solutions need no network during the agent phase.
On the 50 tasks all three harnesses scored, mini-SWE-agent passes 19, OpenCode 20 and Claude Code 20, with every pairwise McNemar $p$ = 1.0, and the flip rates between harnesses are 10 to 26\%.

\end{document}